\documentclass[Afour,sageh,times]{sagej}

\usepackage{moreverb,url}

\usepackage[colorlinks,bookmarksopen,bookmarksnumbered,allcolors=black]{hyperref}

\usepackage{times}
\usepackage{breakcites}
\usepackage{xspace}
\usepackage{amsmath,amssymb,amsfonts,amsfonts,amsthm}
\usepackage{xcolor}
\usepackage{algorithm}
\usepackage[noend]{algorithmic}
\usepackage{graphicx}
\usepackage{listings}	%used for code listings
\usepackage{setspace}
\usepackage{fancyhdr}
\usepackage[nice]{nicefrac}
\usepackage{latexsym}
\usepackage{wrapfig}
\usepackage{colordvi,alltt}
\usepackage{multirow}
\usepackage{tabularx,threeparttable}
\usepackage{fancybox}
\usepackage{calc}
\usepackage{url}
\usepackage{adjustbox}

\usepackage[caption=false]{subfig}
\usepackage{esvect}
\usepackage{booktabs} % Allows the use of \toprule, \midrule and \bottomrule in tables
\usepackage{lipsum}
\usepackage[capitalise,noabbrev]{cleveref}
\usepackage[titletoc,toc,title]{appendix}
\usepackage[utf8]{inputenc}

\usepackage[inline, shortlabels]{enumitem}
\setlist{itemjoin ={,\enspace},itemjoin* = { and\enspace}}
\newlength{\ccBaseWidth}
\newcounter{inlineenum}
\renewcommand{\theinlineenum}{\alph{inlineenum}}

\newcommand\BibTeX{{\rmfamily B\kern-.05em \textsc{i\kern-.025em b}\kern-.08em
T\kern-.1667em\lower.7ex\hbox{E}\kern-.125emX}}

\def\volumeyear{2017}

\newcommand{\ignore}[1]{}

\newcommand{\ompl}{\textsc{OMPL}\xspace}

\newcommand{\pyplot}{Matplotlib\xspace}

\newcommand{\Cpp}{C\raise.08ex\hbox{\tt ++}\xspace}

\newcommand{\R}{\mathbb{R}}

\newcommand{\XX}{\mathbb{X}}
\newcommand{\CC}{\mathbb{C}}

\newcommand{\eps}{\varepsilon}

\DeclareMathOperator{\Ima}{Im}

\newcommand{\dd}{\mathrm{d}\!}
\newcommand{\dds}{\mathrm{d}}

\newcommand{\Simm}{\mathcal{S}}
\newcommand{\EC}{\mathrm{EC}}

\newcommand{\Cs}{\ensuremath{\calC}-space\xspace}

\newcommand{\SLT}{\ensuremath{\Sigma L_2}\xspace}
\newcommand{\Linf}{\ensuremath{\max L_2}\xspace}
\newcommand{\epsinf}{\ensuremath{\eps_\infty}\xspace}
\newcommand{\epst}{\ensuremath{\eps_2}\xspace}
\newcommand{\dk}{\ensuremath{\mathrm{d}_\mathcal{K}}\xspace}

\newcommand{\DOFs}{DOFs\xspace}

\newcommand{\centr}{\textup{Ctd}\xspace}
\newcommand{\tunnelcl}{Permutations}
\newcommand{\chamberscl}{Partitions}
\newcommand{\puzzlecl}{Pebbles}

\newcommand{\etal}{{et~al.}\xspace}

\newcommand{\calA}{\ensuremath{\mathcal{A}}\xspace}
\newcommand{\calC}{\ensuremath{\mathcal{C}}\xspace}
\newcommand{\calD}{\ensuremath{\mathcal{D}}\xspace}

\newcommand{\calK}{\ensuremath{\mathcal{K}}\xspace}

\newcommand{\calT}{\ensuremath{\mathcal{T}}\xspace}
\newcommand{\calU}{\ensuremath{\mathcal{U}}\xspace}

\newcommand{\calX}{\ensuremath{\mathcal{X}}\xspace}

\definecolor{orange}{rgb}{1,0.4,0}

\newcommand{\minisection}[1]{\noindent \textbf{#1.}}
\newcommand{\fcap}{ }
\newcommand{\tcap}{\normalfont\footnotesize}

\newtheorem{thm}{Theorem}[section]

\newtheorem{definition}[thm]{Definition}

  \renewenvironment{thebibliography}[1]{%
    \begin{oldthebibliography}{#1}%
      \footnotesize
      \setlength{\parskip}{0ex}%
      \setlength{\itemsep}{0ex}%
  }%
  {%
    \end{oldthebibliography}%
  }
\def\mypart#1#2{%
  \par\break % Page break
  \vskip .7\vsize % Vertical shift
 ~\refstepcounter{part}% Next part
  {\centering\Huge Part \thepart.\par \\
   \centering #1  
  }% 
  \vskip .1\vsize % Vertical shift 
  
  #2
  \vfill\break % Fill the end of page and page break
}

\def\mypart#1#2
{
	\par\newpage\clearpage % Page break 
	\vspace*{5cm} % Vertical shift 
	\refstepcounter{part}% Next part 
	{
		\centering 
		\textbf
		{
			\Huge Part \thepart
		}
		\par
	}
	\vspace{1cm}% Vertical shift 
	{
		\centering 
		\textbf
		{
			\Huge #1
		}
		\par
	} 
	\vspace{4cm}% Vertical shift 
	#2 
	\vfill
	\pagebreak % Fill the end of page and page break 
}
\def\arxiv{1}

\begin{document}

\runninghead{Atias~\etal}

\title{Effective Metrics for Multi-Robot Motion-Planning}

\author{Aviel Atias, Kiril Solovey, Oren Salzman and Dan Halperin}
%Alistair Smith\affilnum{1} and Hendrik Wittkopf\affilnum{2}}

%\affiliation{\affilnum{1}Sunrise Setting Ltd, UK\\
%\affilnum{2}SAGE Publications Ltd, UK}

%\corrauth{Alistair Smith, Sunrise Setting Ltd
%Brixham Laboratory,
%Freshwater Quarry,
%Brixham, Devon,
%TQ5~8BA, UK.}

%\email{alistair.smith@sunrise-setting.co.uk}
\email{
\mbox{avielatias@gmail.com},
\mbox{kirilsol@post.tau.ac.il},
\mbox{osalzman@andrew.cmu.edu},
\mbox{danha@post.tau.ac.il}}

\begin{abstract}
  We study the effectiveness of metrics for Multi-Robot
  Motion-Planning (MRMP) when using RRT-style sampling-based
  planners. These metrics play the crucial role of determining the
  nearest neighbors of configurations and in that they regulate the
  connectivity of the underlying roadmaps produced by the planners and
  other properties like the quality of solution paths. After screening
  over a dozen different metrics we focus on the five most promising
  ones---two more traditional metrics, and three novel ones which we
  propose here, adapted from the domain of shape-matching.  In
  addition to the novel multi-robot metrics, a central contribution of
  this work are tools to analyze and predict the effectiveness of
  metrics in the MRMP context. We identify a suite of possible
  substructures in the configuration space, for which it is fairly
  easy (i)~to define a so-called \emph{natural distance}, which allows us to predict
  the performance of a metric. This is done by comparing the distribution of its
  values for sampled pairs of configurations to the distribution
  induced by the natural distance; (ii)~to define equivalence classes
  of configurations and test how well a metric covers the different
  classes. We provide experiments that attest to the ability of our
  tools to predict the effectiveness of metrics: those metrics that
  qualify in the analysis yield higher success rate of the planner
  with fewer vertices in the roadmap. We also show how combining
  several metrics together leads to better results (success rate and
  size of roadmap) than using a single metric.

\end{abstract}
%%% Local Variables:
%%% mode: latex
%%% TeX-master: "../paper.tex"
%%% End:


\maketitle

\section{Introduction}\noindent
\begin{figure}[b]
	\begin{center}
        \includegraphics[width=0.5\columnwidth]{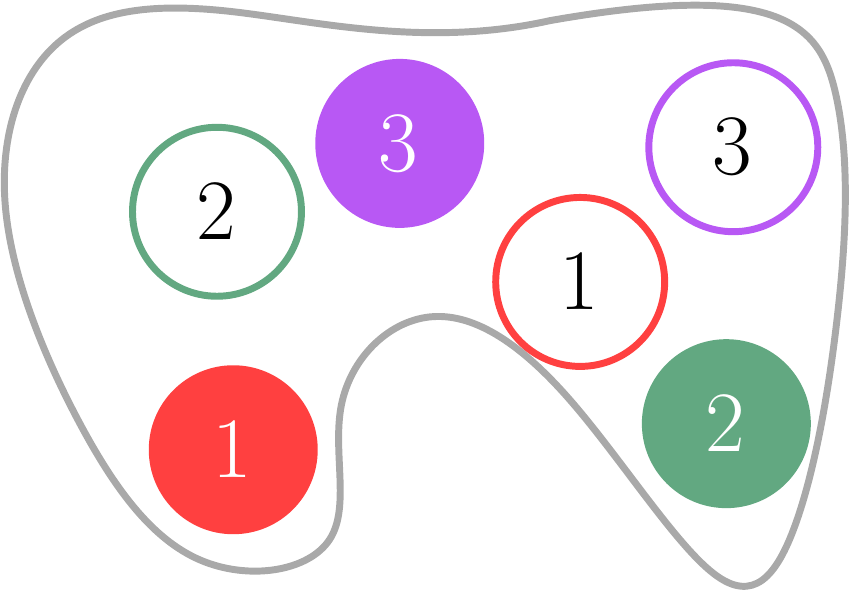}
    \end{center}
    \caption{\fcap An example of an MRMP instance with $m=3$ translating discs. The full discs represent the start configurations and the empty discs represent the goal configurations. Each
disc needs to move from its start configuration to its goal
configuration without hitting the boundary of the workspace
nor its fellow robots.}
    \label{fig:basic_mrmp}
\end{figure}
\emph{Multi-robot motion-planning} (MRMP) is the problem of planning
the motion of a fleet of robots from given start to goal
configurations, while avoiding collisions with obstacles and with each
other. See \cref{fig:basic_mrmp} for a simple illustration. It is a natural extension of the standard single-robot
motion-planning problem. MRMP is notoriously challenging, both from
the theoretical and practical standpoint, as it entails a
prohibitively-large search space, which accounts for a multitude of
robot-obstacle and robot-robot interactions.

Sampling-based planners have proven to be effective in
challenging settings of the single-robot case, and a number of such
planners have been proposed for MRMP~\citep{DobETAL17, 
  DBLP:journals/ijrr/SoloveySH16, DBLP:journals/ras/SvestkaO98,
  DBLP:journals/ai/WagnerC15}. Sampling-based planners attempt to
capture the connectivity of the free space by sampling random
configurations and connecting \emph{nearby} configurations by simple
collision-free paths. In order to measure similarity, or
``closeness'', between a given pair of configurations a \emph{metric}
is employed by the algorithm. The choice of metric has a tremendous
effect on the performance of planners and the quality of the returned
solutions (see \cref{sec:related} for further discussion about metrics
that are tailored for various robotic systems). Nevertheless, no
specialized metrics for multi-robot systems have been proposed, to the
best of our knowledge.

\begin{figure}
    \centering
	\subfloat[\fcap ``Easy'' to-connect configurations]
	{
		\includegraphics[width=0.4\columnwidth]{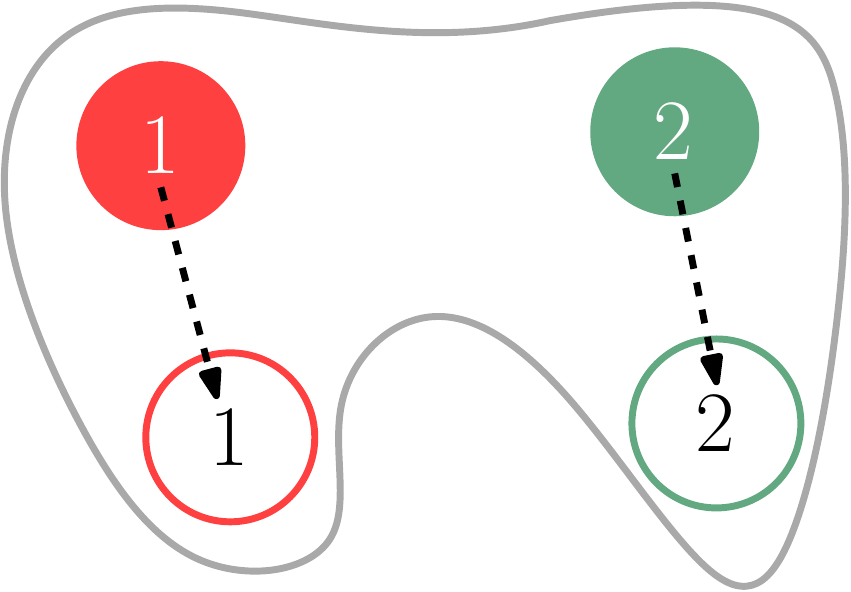}
		\label{subfig:motivation_easy}
	}
	\qquad
	\subfloat[\fcap ``Hard'' to-connect configurations]
	{
		\includegraphics[width=0.4\columnwidth]{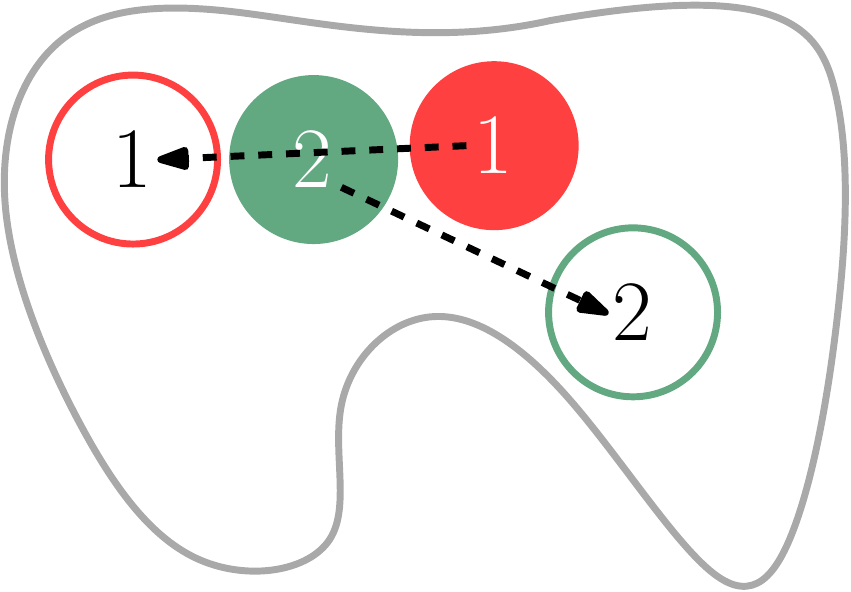}
		\label{subfig:motivation_hard}
	}
	
    \caption{\fcap Example for the setting of two disc robots, drawn in red and green, respectively. Full discs represent start configurations, and empty discs represent goal configurations. Since it is easier to connect the configurations in~(a) when compared to the configurations in~(b), the distance in~(a) should be \textbf{smaller} than the distance in~(b).}
    \label{fig:motivation}
\end{figure}

\begin{figure}[t]
  \centering
  \includegraphics[width=0.9\columnwidth]{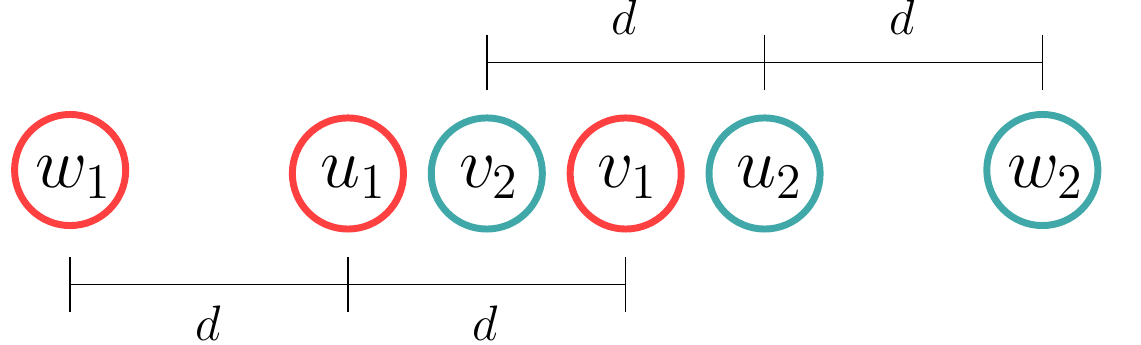}
  \caption{\fcap Example of \SLT for the setting of $m=2$ disc robots in the
    plane. The red discs, centered in $u_1,v_1,w_1$ represent possible
    positions for the first robot, whereas the green discs, centered in
    $u_2,v_2,w_2$, represent possible positions for the second
    robot. We set the positions in the following manner:
    $\|u_1-v_1\|_2=\|u_1-w_1\|_2, \|u_2-v_2\|_2=\|u_2-w_2\|_2$.
    $U=(u_1,u_2),V=(v_1,v_2),W=(w_1,w_2)$ represent three
    simultaneous placements of the two robots.  While
    $\SLT \left(U,V\right)=\SLT\left(U,W\right)$, it is intuitive that
    it is easier to connect $U$ to $W$ rather than to $V$. This
    example hints that \SLT may not be suitable for all cases as it fails to capture robot-robot interaction.}
  \label{figure:basic_l2}
\end{figure}

A common approach (see ~\citep[pp. 210]{clhbkt05}) states that the metric should reflect how difficult it is to plan a path between two configurations. See \cref{fig:motivation} for an illustration.
Nowadays, a common metric for multi-robot systems is defined as a sum
of metric values for single robots
(\citep{DBLP:conf/wafr/PlakuK06,DBLP:journals/ijrr/SoloveySH16}, and
in fact this is the default in
\ompl~(Open Motion Planning Library)~\citep{DBLP:journals/ram/SucanMK12}), i.e., the sum of distances
induced by each of the robots separately. We denote this metric by
\SLT (to be formally defined in \cref{sec:metrics}). This metric does
not always adequately express distance in the \emph{configuration
  space} (\Cs) because it does not account for interactions between
different robots. A simple example is shown in \cref{figure:basic_l2}.

\subsection{Contribution}\noindent
%In this work we consider the problem of devising good distance metrics
%for MRMP.  We proceed to introduce several new metrics for MRMP and
%show that they improve upon the standard \SLT metric in various
%settings.  Our new metrics combine ideas from various fields of study
%such as computational geometry, shape matching and image
%processing. The main benefit of the new metrics, is that they do not
%only take the relative positions of the same robot into consideration,
%but also the interactions between the different robots.  We consider
%several properties that such metrics should maintain and describe how
%to analyze those properties for a given metric.
%
%We present experimental results, which show that our metrics improve
%the effectiveness of motion planning, even in complex
%environments, and suggest to use this type of metrics side-by-side
%with traditional multi-robot metrics in order to be able to
%effectively solve various problem instances. 

  We study the effectiveness of metrics for MRMP when using
  RRT-style sampling-based planners. These metrics play the
  crucial role of determining the nearest neighbors of
  configurations and in that they regulate the connectivity of
  the underlying roadmaps produced by the planners and other
  properties like the quality of solution paths. After screening
  over a dozen different metrics we focus on the five most
  promising ones---two more traditional metrics, and three novel
  ones which we propose here, adapted from the domain of
  shape-matching.  In addition to the novel multi-robot metrics,
  a central contribution of this work are tools to analyze and
  predict the effectiveness of metrics in the MRMP context. We
  identify a suite of possible substructures in the configuration
  space, for which it is fairly easy (i)~to define a so-called
  \emph{natural distance}, which allows us to predict the
  performance of a metric. This is done by comparing the
  distribution of its values for sampled pairs of configurations
  to the distribution induced by the natural distance; (ii)~to
  define equivalence classes of configurations and test how well
  a metric covers the different classes. We provide experiments
  that attest to the ability of our tools to predict the
  effectiveness of metrics: those metrics that qualify in the
  analysis yield higher success rate of the planner with fewer
  vertices in the roadmap. We also show how combining several
  metrics together leads to better results (success rate and size
  of roadmap) than using a single metric.

\subsection{Organization}\noindent
The organization of this paper is as follows.  In \cref{sec:related}
we review related work.  In \cref{sec:screening} we describe the early
phase of our investigation, where we tested a large number of metrics
with different planners, and explain why we chose the metrics and planner
on which we focus in the sequel. In \cref{sec:metrics} we formally define five
metrics which will be discussed later.  In \cref{sec:sub,sec:analysis}
we present methods for analyzing the proposed metrics using
identification of substructures arising in MRMP.  In
\cref{sec:experiments} we provide experimental results allowing us to
compare the utility of the metrics. In \cref{sec:rotate} we extend the metrics for robotic systems other than those we discuss earlier. Finally, in \cref{sec:future} we
outline possible future work.
\ifx\arxiv\undefined
An extended version of this paper, to which we refer throughout the text, is available at \url{https://arxiv.org/abs/1705.10300}.
\else
\fi
%The paper is accompanied with
%supplementary material, to which we refer throughout the text.

%%% Local Variables:
%%% mode: latex
%%% TeX-master: "../ijrr.tex"
%%% End:

\section{Related Work}
\label{sec:related}\noindent
We start this section with work related to multi-robot motion planning
(MRMP). Then, we proceed to discuss metrics in the context of robotics
and beyond.  We assume some familiarity with basic concepts of
sampling-based motion planning (see, e.g.,~\citet{clhbkt05,HKS16,lavalle06}).

\subsection{Multi-robot motion-planning}\noindent
Initial work on MRMP
focused on exact methods for solving the problem instance.
\citet{schwartz1983piano} consider the case of disc robots
operating in an environment cluttered with polygonal
obstacles. Their algorithm runs in time polynomial in the
obstacles' complexity, but exponential in the number of robots.
Several works address the case when the number of robots is
bounded.  \citet{aronov1999motion} present a technique that
reduces the complexity of the problem for two or three robots.% ,
% making it feasible in the case that the obstacles are
% sparse. Kiril: this sentence was ambiguous. 
\citet{DBLP:journals/amai/SharirS91} propose an approach to
coordinate the motion between two robots of various types, i.e., not necessarily translating robots.

\citet{hss-cmpmio} and \citet{hopcroft1986reducing} prove that MRMP is PSPACE-hard even when the robots are rectangular and operate in a rectangular region. Later, \citet{DBLP:journals/tcs/HearnD05,DBLP:books/daglib/0023750} extend the result to rectangles of size $1\times 2$ and $2\times 1$. The problem is strongly NP-hard also when the robots are translating discs~\citep{SY84}. The proof in \citep{SY84} makes use of robots that differ in their size. In a recent result~\citep{DBLP:journals/ijrr/SoloveyH16}, the setting of unit-square robots and polygonal obstacles is considered. The problem is proven to be PSPACE-hard. The result holds even in case that all the robots are identical and indistinguishable (namely, in the \emph{unlabeled} setting). In addition to the aforementioned results, system dynamics can introduce additional complications to the problem~\citep{Johnson-RSS-16}.

The unlabeled variant of the problem was
introduced by~\citet{kh-pim05} who describe a sampling-based
planner for the problem. Although this problem is hard in
general, under some simplifying assumptions it can be solved in
polynomial time as function of the number of robots and the
complexity of the workspace environment. For disc-shape robots,
under some assumptions on the free space and the separation
between initial and goal configurations, it is possible to find a
solution in polynomial time~\citep{DBLP:journals/arobots/TurpinMMK14}.
Furthermore, the obtained solution is
optimal with respect to the longest distance traveled by any one
robot. \citet{DBLP:journals/tase/AdlerBHS15} describe a more
efficient algorithm, which guarantees to find a solution, but not
necessarily the best one.  Using similar conditions a nearly
optimal solution (with respect to the sum of path lengths) can be found
in polynomial time~\citep{SolETAL15}. Finally, we mention that
the $k$-\emph{color} generalization, where the
robots are partitioned into $k$ groups and the robots in each
group are indistinguishable, has been studied using a sampling-based approach~\citep{sh-kcmr}.

Approaches for solving MRMP can be roughly subdivided into two types:
\emph{coupled} and \emph{decoupled}.  In the latter approach (see,
e.g.,~\citet{DBLP:journals/ijrr/BareissB15,lls-mpcmr,bo-pmpmr}), a path
or an initial plan are found for each robot separately, and then the
paths are coordinated with each other.
%Although this approach can cope with larger fleets of robots in some settings, it gives no completeness guarantees.
Although this approach is less sensitive to the number of robots, when compared with the coupled approach, it gives no completeness guarantees. 

The coupled approach usually treats the entire system as a single
robot, for which the number of \emph{degrees of freedom} (\DOFs) is
equal to the sum of the number of \DOFs of the individual robots in
the system. This approach usually comes with stronger theoretical
guarantees such as completeness~\citep{kh-pim05, shh-mms2, sl-dcc,
  sh-kcmr, DBLP:journals/ijrr/SoloveySH16} or even optimality~\citep{DBLP:journals/ai/WagnerC15,DobETAL17} of the returned solutions. However,
due to the computational hardness of MRMP~\citep{DBLP:journals/tcs/HearnD05, hss-cmpmio, Johnson-RSS-16,
  DBLP:journals/ijrr/SoloveyH16, SY84}, coupled techniques
do not scale well with the increase in the number of robots. 
% We do mention that, when simplifying assumptions are made concerning the separation of initial and goal positions, MRMP can be solved in polynomial time, as function of the number of robots and the complexity of the workspace environment (see, \cite{DBLP:journals/tase/AdlerBHS15, SolETAL15, DBLP:conf/icra/TurpinMK13}).

\subsection{Metrics}\noindent
% basic metrics researches (Amato, Kuffner)
The choice of a metric for nearest-neighbors queries in a
sampling-based planner can be
crucial. \citet{DBLP:journals/trob/AmatoBDJV00} were the first to study
the effect of a metric on sampling-based planners. 
They consider PRM~\citep{DBLP:journals/trob/KavrakiSLO96} as the planner and define effectiveness as the number of discovered edges in the roadmap.
They compare effectiveness of some variants of the Euclidean metric in settings that involve translation and rotation of a single robot.
%\citeauthor{DBLP:conf/icra/Kuffner04}~
\citet{DBLP:conf/icra/Kuffner04} considers metrics for
rigid-body motion and proposes an interpolation between the rotation
component and the translation component.

% kinodynamic metrics
Extensive research
has been carried out in order to find suitable metrics for
other settings of motion planning, such as robots with differential
constraints~\citep{DBLP:conf/iros/BharatheeshaCWW14,
  DBLP:conf/iros/BoeufCAS15, DBLP:conf/icra/LaValleK99,
  DBLP:conf/icra/PalmieriA15}.

% multirobot
\citet{DBLP:journals/trob/PamechaEC97} analyze metrics for systems
with a single robot consisting of multiple modules that must stay in
touch with each other (\emph{multi-module systems}). Though any
module can be thought of as a robot, the system restrictions are that
modules are only allowed to move on a grid, and must stay in contact
in order to form a metamorphic robot. Hence, their results are not
straightforward to extend to arbitrary multi-robot systems. 
Further analysis for multi-module systems can be found in \citet{DBLP:conf/robio/WinklerWF11} and \citet{DBLP:journals/trob/ZykovMDL07}.

% learning, mixing
Recent methods employ machine learning to develop metrics that are
tailored to the specific motion-planning problem at hand. \citet{DBLP:conf/iros/EkennaJTA13} introduce a framework in
which there is a candidate set of metrics, and the planner adaptively
selects a metric on-the-fly. The selection may vary over time or
between different regions of the workspace.  This implies that a set
of metrics, each suitable for a different setting, can be combined in
order to solve more diverse settings that consist of smaller,
specific, (sub)settings.  \citet{DBLP:conf/wafr/MoralesTPRA04} have the same
observation that different portions of the \Cs may behave differently.
In our work we will also refer to the case where different metrics are
more effective than others in different portions of the \Cs.

% shape-matching
Estimating distances between sets of points is in broad use in
shape matching (see the survey~
\citep{DBLP:series/acvpr/VeltkampH01}).  Such techniques (see,
e.g.,~\citet{DBLP:journals/pami/BelongieMP02}) are concerned with
estimating the distance between shapes and with finding a matching
between shapes.  \citet{Kendall} provides a rigorous mathematical
study of the subject, where point sets are mapped to high-dimensional
points, on which distance measures can be more easily defined (see
more details in~\cref{sec:metrics}).

% graph drawing
Another area where distance between sets of points is of interest is
graph drawing. \mbox{\citet{DBLP:journals/jgaa/BridgemanT00}} list a
large number of distance metrics between planar graphs. Some of the
metrics give a significant weight to the relative order between the
nodes, which is also the guideline for the metrics we propose in this
paper. \mbox{\citet{DBLP:journals/jgaa/LyonsMR98}} address the same
problem, and measure similarity based on both Euclidean distance and
relative order between the nodes.

%%% Local Variables:
%%% mode: latex
%%% TeX-master: "../paper.tex"
%%% End:

\section{Initial Screening}
\label{sec:screening}
\noindent We began our study by experimenting with four different
planners, fifteen different metrics and variations of them.  For
planners we tried RRT-style and
EST-style~\citep{DBLP:journals/ijcga/HsuLM99} planners that are
adapted to the multi-robot setting. We tested both single-tree and
bi-directional variants of each algorithm. PRM-style planners cannot
cope with the induced high-dimensional space.  RRT-style planners
showed much better success rate in solving MRMP problems when compared
to EST-style planners.  This is why the study continues henceforth
with dRRT~\citep{DBLP:journals/ijrr/SoloveySH16}---an adaptation of
RRT to the multi-robot setting, which can cope with a larger number
of robots and more complicated tasks than RRT as-is. We mention that
M*~\citep{DBLP:journals/ai/WagnerC15}, which is another sampling-based planner tailored for MRMP, is less
relevant to our current discussion since it only employs metrics
concerning individual robots.

For metrics, we began by following the common approach of choosing
metrics that have high correlation with the failure rate of the local
planner~\citep[pp. 210]{clhbkt05}. Note that this is also the
guideline behind using the swept volume and its approximations as a
metric for rotating
robots~\citep{DBLP:journals/trob/AmatoBDJV00,DBLP:conf/iros/EkennaJTA13,DBLP:conf/icra/Kuffner04}. It
turns out that when using such metrics with RRT-style planners, the
exploration of the \Cs is unbalanced--- the explored configurations
tend to have the robots separated from each other.  The analogue for
single-robot planning is exploration of configurations that tend to be
far from obstacles, avoiding paths that go near the obstacles. This
phenomenon is further discussed in 
\ifx\arxiv\undefined
the extended version of this paper.
\else
Appendix~\ref{supp:visualization}.
\fi

We continue with metrics that adapt geometric methods from the domain
of
shape-matching~\citep{DBLP:journals/pami/BelongieMP02,DBLP:journals/jct/Goodman80,DBLP:journals/siamcomp/GoodmanP83,Kendall},
including existing methods that are used for mismatch
measure~\citep{DBLP:journals/dcg/AltMWW88}.  We also used measures of
similarities that are employed in the domain of
graph-drawing~\citep{DBLP:journals/jgaa/BridgemanT00,DBLP:journals/jgaa/LyonsMR98}.

Out of the fifteen tested metrics and their variations, we remained
with the five most successful metrics that are described below in
\cref{sec:metrics}.

% \aviel{Please review the following paragraph. Old version is:}
% Finally, we mention that we experimented with several types of robots
% including planar ones that are allowed to translate and
% rotate. However, we chose to conduct our final experiments with robots
% bound to translate in the plane, as it makes the presentation
% clearer. Moreover, we believe that the study of complex rigid-body
% motion~\cite{DBLP:conf/icra/Kuffner04} in the context of metrics is
% mostly orthogonal to our current efforts.

Finally,
we mention that we experimented with several types of robots
including planar ones that are allowed to translate and
rotate. All the metrics in this paper can cope with such robots
(see \cref{sec:rotate}). However, we chose to conduct our final
experiments and analyses with robots bound to translate in the
plane, as it makes the presentation clearer. Moreover, we believe
that the study of complex rigid-body
motion~\citep{DBLP:conf/icra/Kuffner04} in the context of metrics
is mostly orthogonal to our current efforts of incorporating
multi-robot considerations into the metric. Robotic systems
  involving dynamics are outside the scope of this paper, and we
  leave their study for future work.

%%% Local Variables:
%%% mode: latex
%%% TeX-master: "../paper.tex"
%%% End:

\section{Metrics for Multi-Robot Motion-Planning}\label{sec:metrics}
\noindent In this section we discuss the role of metrics in
sampling-based MRMP. Then, we formally define the standard
\SLT, \Linf metrics and introduce the metrics \epst, \epsinf, \centr,
which will be evaluated in \cref{sec:experiments}.

% \reviewer{We should ``better motivate where translation only (disc) robots are useful and prevelant''.} \oren{rumbas, amazon robotics can be roughly approximated as discs etc.}

% \reviewer{Some of the distance metric requirements (symmetry, triangle inequality) do not work when systems also have dynamics. The authors should address this in their
% dicussion.} \aviel{We just need to address it and say it is outside of the scope of this paper.}

We consider $m$ robots $r_1,\ldots,r_{m}$ operating in a shared
workspace. For simplicity we assume that the robots are identical
in shape and function. The \Cs of each individual robot can
be denoted by some $\calX$. Note that we still distinguish
between the different robots. We assume that each $r_i$
represents a translating disc in the plane, and so $\calX=\R^2$.
Denote the \emph{joint \Cs} for the $m$ individual robots by
$\calX^m=\calX\times \ldots \times \calX$, i.e., a \emph{joint
  configuration} $U=(u_1,\ldots,u_m)$ represents a set of
configurations for the $m$ robots.

Our presentation focuses on translating disc robots, which are 
  often encountered in practice as iRobots, TurtleBots, or as the bounding
  volume to more complex systems. However, we note that the metrics
  described below can be extended to more general settings of the
  problem, such as non-disc robots and 3D environments. See
  \cref{sec:rotate} for more details.
  % Robotic systems involving dynamics are outside the scope of this paper, and we defer it for future work.}

Sampling-based tools for single and multi-robot systems rely on
metrics to measure similarity between configurations. Let $U,V,W$ be
joint configurations of our multi-robot system.  A metric in the
context of MRMP is a distance function
$\dd:\calX^m \times \calX^m \rightarrow \left[ 0, \infty \right)$,
which satisfies the five properties:
\begin{enumerate}[(a) ]% 
\item \label{metric:positive} \emph{non-negativity}: $\dd \left( U, V \right) \geq 0$;
\item \label{metric:identity1} \emph{identity}: $\dd\left(U,U\right)=0$;
\item \label{metric:identity2} \emph{identity of indiscernibles}: $\dd\left(U,V\right)=\!0 \Rightarrow U=V$;
\item \label{metric:symmetric} \emph{symmetry}: $\dd \left( U, V \right)=\dd \left(V,U\right)$;
\item \label{metric:triangle} \emph{triangle inequality}: $\dd \left(U,W\right) \leq \dd \left(U,V\right) + \dd \left(V,W\right)$.
\end{enumerate}
Efficient nearest-neighbors data structures usually do not rely on
property~\ref{metric:identity2}~(see, e.g.,~\citep{DBLP:conf/vldb/Brin95,DBLP:journals/csur/ChavezNBM01,DBLP:conf/vldb/CiacciaPZ97}),
and so can be applied to \emph{pseudometrics}, which satisfy
properties~\ref{metric:positive}, \ref{metric:identity1},
\ref{metric:symmetric} and~\ref{metric:triangle}.  We extend the
discussion also to \emph{pseudosemimetrics} which are functions that
satisfy only properties~\ref{metric:positive}, \ref{metric:identity1},
and ~\ref{metric:symmetric}. In that case, we cannot use sophisticated
data structures that rely on the triangle inequality.  For
simplicity, from now on we will refer to any
pseudosemimetric as a \emph{metric}. \vspace{5pt}

% \reviewer{It seems that epsiloncongruence will not work well when
%   robots become very far apart.} \aviel{We implicitly respond to
%   this in the comment about $\tau$. Do you think it is enough?} \kiril{Yes.}

\minisection{Standard metrics} The following two metrics are simple
extensions of single-robot metrics to the multi-robot setting.  Let
$L$ be a single-robot metric
$L:\calX\times\calX\rightarrow \left[ 0, \infty \right)$.  For any two
joint configurations
$U=\left(u_1,\ldots,u_m\right),V=\left(v_1,\ldots,v_m\right) \in
\calX^m$ we define $\Sigma L$ and $\max L$ as:
\begin{align*}
\Sigma L\left(U,V\right)  &= \sum_{i=1}^m L\left(u_i,v_i\right), \\
\max L\left(U,V\right)  &= \max_{i=1,\ldots,m} L\left(u_i,v_i\right).
\end{align*}

%A standard case is where $L$ is the standard Euclidean distance, denoted by $L_2$. %; for any two single-robot configurations $u=\left(u_x,u_y\right),v=\left(v_x,v_y\right)\in\calX$ the Euclidean distance in $\RR^2$. %is:
%$$
%L_2 \left(u,v\right)=\sqrt{\left(u_x-v_x\right)^2 +
%  \left(u_y-v_y\right)^2}.
%$$
We consider the two metrics obtained by setting $L=L_2$, which is the standard Euclidean distance, and denote
them by \SLT and \Linf.  Those metrics satisfy
properties~\ref{metric:positive}-\ref{metric:triangle}. We note that
the former is used by default in many settings, whereas the latter has
earned much less attention.  \vspace{5pt}

\minisection{$\eps$-congruence metrics} Here we introduce new metrics,
which are based on the notion of \emph{approximate congruence} or
\emph{$\eps$-congruence}, described by
\citet{DBLP:journals/dcg/AltMWW88}.

\begin{definition}[\emph{$\eps$-congruence}]
  Let $L:\mathcal{X} \times \mathcal{X} \rightarrow [0,\infty) $ be a
  single-robot metric, and let $\calT$ be the set of all translations
  $T:\calX\rightarrow \calX$. For every two joint configurations
  $U=\left(u_1,\ldots,u_m\right),V=\left(v_1,\ldots,v_m\right) \in
  \calX^m$ the $\eps$-congruence with respect to $L$ is defined as
  \[ \eps_L \left(U,V\right) = \min_{T\in\calT} \max_{i=1,\ldots,m}
  L\left( T\left(u_i\right),v_i\right). \]
\end{definition}

This metric expresses the required tolerance (with respect to
$L$) for the two sets of points to be equivalent to each other under
translation.
% \begin{wrapfigure}{r}{0.25\textwidth} \vspace{-10pt}
%   \centering
%   \includegraphics[width=0.25\textwidth, clip, trim=71pt 50pt 57pt 11pt]{figures_basic_epsilon2_13.pdf}
%  \caption{\fcap $\eps$-congruence with respect to $L_2$.
%  	  $U$ is marked with circles, $V$ with
%      squares, and the translated configuration $T\left(U\right)$ with stars.
%      Each of the $m=5$ robots is denoted by a different color.
%      If each star falls inside its corresponding ball then the 
%      balls' (common) radius corresponds to a valid translation. 
%      The $\eps$-congruence is the minimal valid radius.}
%     \label{fig:eps_congruence}
%     \reduce \reduce
% \end{wrapfigure}

\begin{figure}[t]
  \centering
  \includegraphics[width=0.8\columnwidth,trim={80pt 40pt 60pt 0},clip]{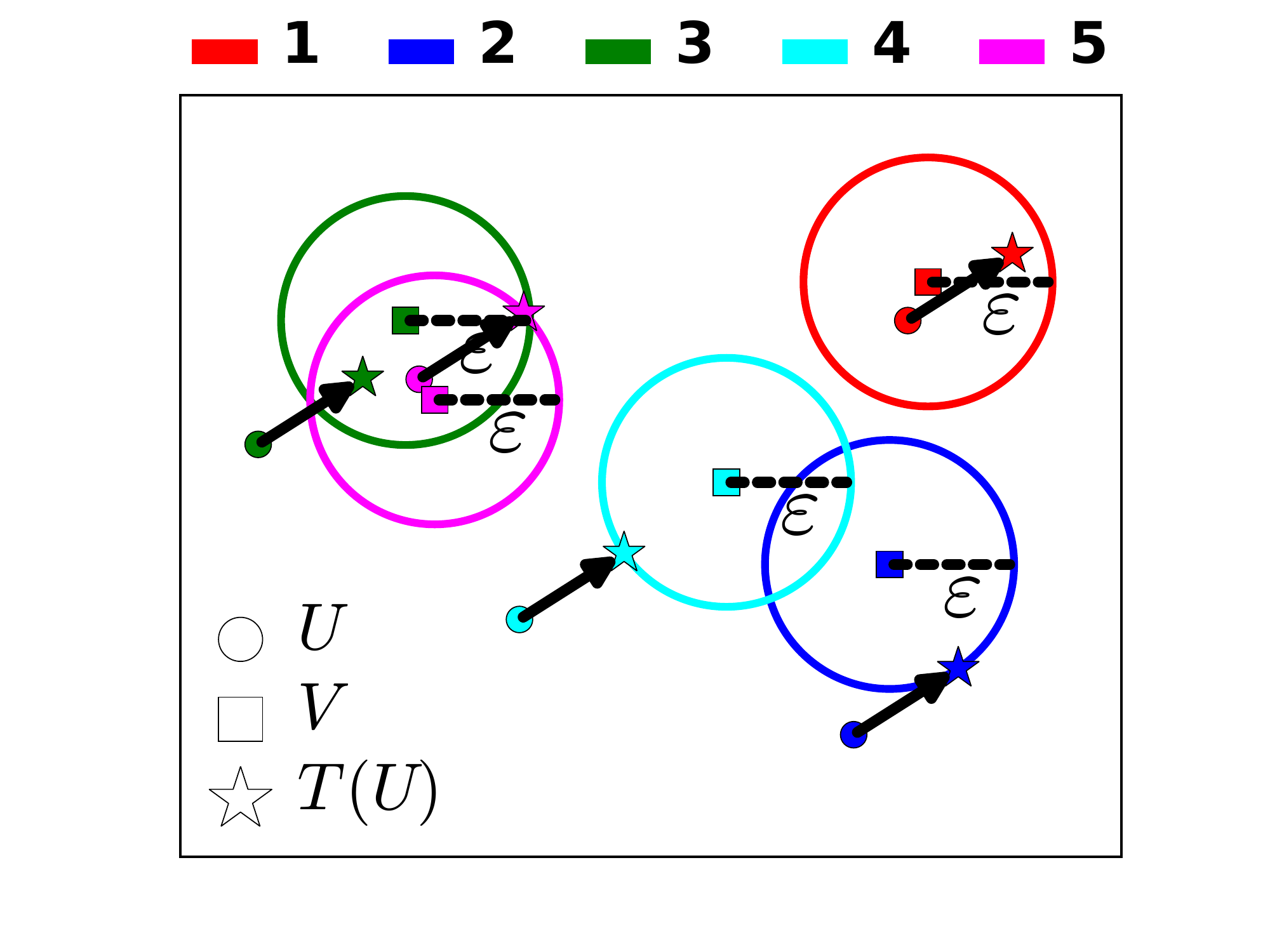}
  \caption{\fcap $\eps$-congruence with respect to $L_2$.
  	  $U$ is marked with circles, $V$ with
      squares, and the translated configuration $T\left(U\right)$ with stars.
      Each of the $m=5$ robots is denoted by a different color.
      If each star falls inside its corresponding ball then the 
      balls' (common) radius corresponds to a valid translation. 
      The $\eps$-congruence is the minimal valid radius.}
    \label{fig:eps_congruence}
\end{figure}

We denote $\eps$-congruence with respect to $L_2$ and $L_\infty$ by
\epst and \epsinf, respectively. See an illustration
% on the right: $U$ is
%marked with circles, $V$ with squares, and the translated
%configuration $T\left(U\right)$ with stars.  Each of the $m=5$ robots
%is denoted by a different color.  If each star falls inside its
%corresponding ball then the balls' (common) radius corresponds to a
%valid translation.  The $\eps$-congruence is the minimal valid radius.
in \cref{fig:eps_congruence}. 

Note that $\eps$-congruence
satisfies all the properties of a pseudosemimetric, and in case $L$
satisfies the triangle inequality (which is the case for $L_2$ and
$L_\infty$) then $\eps$-congruence is a pseudometric and therefore can
be used with any nearest-neighbor data structure.

\vspace{5pt}

\minisection{Shape-based metric}
\ifx\arxiv\undefined
Let $U=\left(u_1,\ldots,u_m\right)$ and $V=\left(v_1,\ldots,v_m\right)$ be two joint configurations for $m$ robots. Denote by $x_i$ and $y_i$ the $x$ and $y$ coordinates (respectively) of $v_i-u_i$.
The \emph{Centroid} distance is defined as the sum of squared Euclidean distances between $v_i-u_i$ and the common centroid of $\left\{v_i-u_i\right\}_{i=1}^m$.
The centroid distance is calculated using the following equation:

\begin{equation}
\label{eq:cetnroid_def}
\centr\left(U,V\right)\!=\! 
\sum_{i=1}^m {\left(x_i^2+y_i^2\right)} - 
\frac{\left( \sum_{i=1}^m x_i\right)^2 + \left(\sum_{i=1}^m y_i\right)^2 }{m}.
\end{equation}
The development of \cref{eq:cetnroid_def} is based on the notion of \emph{shape space}~\citep{Kendall}. Refer to the extended version of this paper for intuition and full details.
\else
To measure the mismatch between two
point sets, \citeauthor{Kendall} introduced the notion of a
\emph{shape space}~\citep{Kendall}.  Specifically, given $m$
$k$-dimensional points the shape space
$\Sigma_k^m = \R^{k\times m} /\Simm$ is the quotient space
of $\R^{k\times m}$ by the group of similarities generated by
translations, rotations and dilations.
Namely, it is a subdivision of all point sets into equivalence
classes, where two point sets are equivalent if one can be transformed to
the other by some operation $T \in \Simm$.

Let
$U=\left(u_1,\ldots,u_m\right), V=\left(v_1,\ldots,v_m\right) \in
\Sigma_k^m$
and \mbox{$T \in\Simm$}.  Note that by the definition of equivalence sets we
have that the distance between $U$ and $V$ is equal to the distance
between $T(U)$ and $V$.  This allows us to define the mismatch between
$U$ and $V$ as the minimal distance over all similarities
$T \in\Simm$.  Specifically, \citeauthor{Kendall} uses the sum of
squares of distances between associated pairs of points.  Thus, the
distance between two point sets is defined as\footnote{$[T(U)]_i$ is the
  $i$th planar point in the vector of $m$ such points $T(U)$.}
\begin{equation}
  \label{eq:shape_distance}
  \min_{T\in\Simm} \left\{ \sum_{i=1}^m \left(
      L_2\left(\left[T\left(U\right)\right]_i,v_i\right) \right)^2 \right\}.
\end{equation}

We propose to adapt these ideas to the setting of MRMP.  Specifically,
in our basic setting we have that
\begin{enumerate*}[(i)]
	\item each single-robot configuration is a planar point in $\R^2$
	\item we restrict the set of similarities $\Simm$ to translations only.
\end{enumerate*}
Thus, we can rewrite \cref{eq:shape_distance} as:
\begin{equation}
\label{eq:translation_distance}
\min_{T\in\R^{2m}} \left\{ \sum_{i=1}^m \left(L_2\left(T_i,v_i-u_i\right)\right)^2 \right\},
\end{equation}
where $T_i$ is the translation component in $T$ of the $i$-th point.

We restrict $\Simm$ to translations only since we are using a local
planner that generates a straight-line path for each robot.
Such local planning between a configuration $U$ and a translation of it $T\left(U\right)$ is always
free of robot-robot collisions.
However, it may not be free of collisions if we allow rotations and dilations.

The translation $T$ that minimizes \cref{eq:translation_distance} is known as the \emph{centroid} of the set $\left\{v_i-u_i\right\}_{i=1}^m$ (see \citep[pp. 520]{protter1977college}).
For two-dimensional points ($k=2$) the minimal value is
%\begin{equation}
%\label{eq:cetnroid_def}
%\centr\left(U,V\right)\!=\! 
%\sum_{i=1}^m {x_i^2} + \sum_{i=1}^m {y_i^2} - 
%\frac{\left( \sum_{i=1}^m x_i\right)^2 + \left(\sum_{i=1}^m y_i\right)^2 }{m},
%\end{equation}
\begin{equation}
\label{eq:cetnroid_def}
\centr\left(U,V\right)\!=\! 
\sum_{i=1}^m {\left(x_i^2+y_i^2\right)} - 
\frac{\left( \sum_{i=1}^m x_i\right)^2 + \left(\sum_{i=1}^m y_i\right)^2 }{m},
\end{equation}
where $x_i$ and $y_i$ are the $x$ and $y$ coordinates of $v_i-u_i$.
\cref{eq:cetnroid_def} defines the \emph{Centroid} distance in 2D,
which we denote by \centr.
\fi
\vspace{5pt}

In summary, we have presented five metrics for MRMP: the more traditional
\SLT and \Linf, and the novel metrics \epst, \epsinf, \centr. We will
evaluate these five metrics below. 

%%% Local Variables:
%%% mode: latex
%%% TeX-master: "../paper"
%%% End:

\section{Canonical Substructures in \boldmath\Cs}\label{sec:sub}\noindent
Here we introduce a new approach to better conquer the intricate
problem of MRMP.  We identify several ``gadgets'', which represent
local instances of the problem, and which force the robots to
coordinate in a specific and prescribed manner. Those gadgets can be
viewed as a set of representative tasks that need to be carried out in
typical scenarios of MRMP. Examining these substructures, rather than
the entire complex problem, has two benefits. Firstly, such
substructures can be straightforwardly decomposed into a small number
of equivalence classes (ECs) of (joint) configurations, which can be viewed
as a discrete summary of the continuous problem. We conjecture that a
metric which maximizes the number of explored ECs by a
given planner also leads to better performance of the planner.
Secondly, those ECs of a given substructure, and the
relations between them, induce a \emph{natural distance metric}, which
faithfully quantifies how difficult it is to move between any given pair
of joint configurations. This gives an additional method to assess
the quality of a given metric by comparing it to the natural metric.

In the remainder of this section we describe three such canonical
substructures, which we refer to as Permutations, Partitions, and
Pebbles, and denote them by
$\XX_{\textup{\tunnelcl}}, \XX_{\textup{\chamberscl}},
\XX_{\textup{\puzzlecl}}$.
We also describe their corresponding natural
metrics. In~\cref{sec:analysis} we describe tools for analysis of
metrics. Of course there could be many more useful substructures---see
comment in the concluding section.

Each such substructure $\XX$ is a subset of the joint \Cs~$\calX^m$.
For every $\XX$ we identify a finite collection of $e>0$ disjoint
subsets $X_1,\ldots,X_e$ of $\XX$ termed \emph{equivalence classes}
(ECs). Note that each EC is a subset of the joint \Cs.  We say that
two joint configurations $U,V\in \XX$ are \emph{equivalent} if they
belong to the same EC $X_i$. If robots can also leave one EC $X_i$ and
enter another $X_{i'}$, without going through any other EC then we say
that the ECs $X_i,X_{i'}$ are \emph{neighbors}. This gives rise to the
\emph{equivalence graph} $G_\XX$ whose vertices are the ECs of $\XX$,
and there is an edge between every two neighboring ECs.

We are now ready to define the \emph{natural distance} \dk between two
given joint configurations $U,V\in \XX$. For a given $U\in \XX$ denote
by $\EC\left(U\right)$ the EC of $\XX$ in which it resides. Then the
natural distance $\dk\left(U,V\right)$ is the graph distance over
$G_\XX$ between $\EC(U)$ and $\EC(V)$, namely the number of edges
along the shortest path in the graph between the vertices
corresponding to $\EC(U), \EC(V)$.

\subsection{\tunnelcl} 
\begin{figure}
    \centering
	\subfloat[\fcap]
	{
		\includegraphics[width=0.231\textwidth]{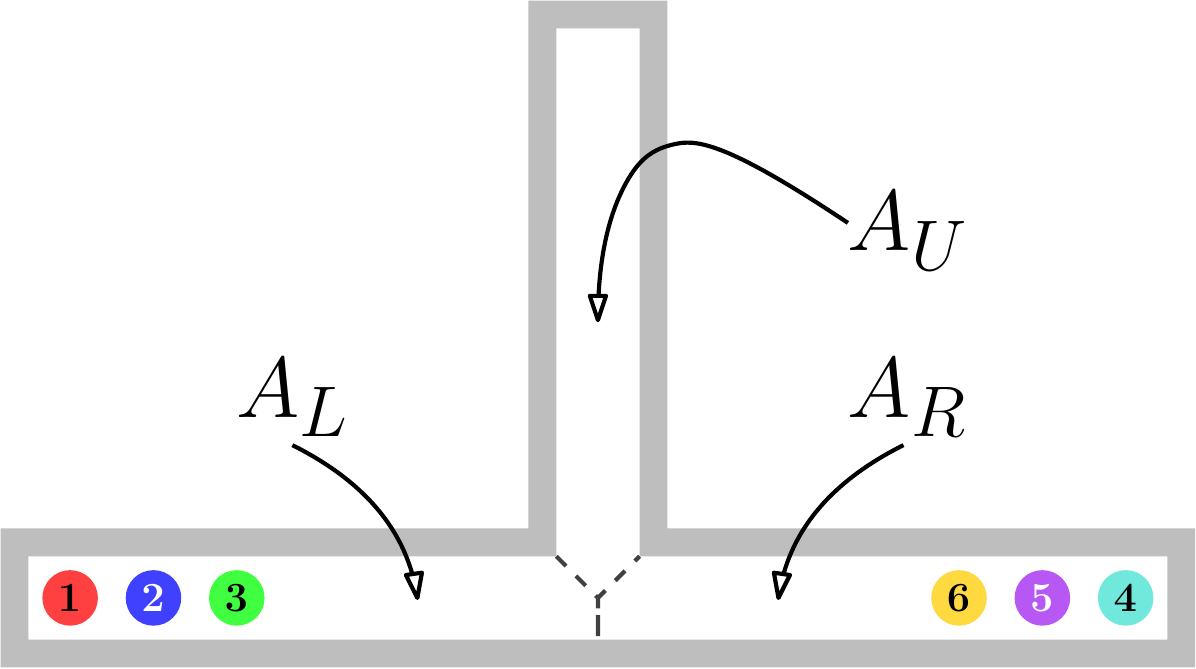}
		\label{subfig:tunnel_env_start}
	}
	%\qquad
	\subfloat[\fcap]
	{
		\includegraphics[width=0.231\textwidth]{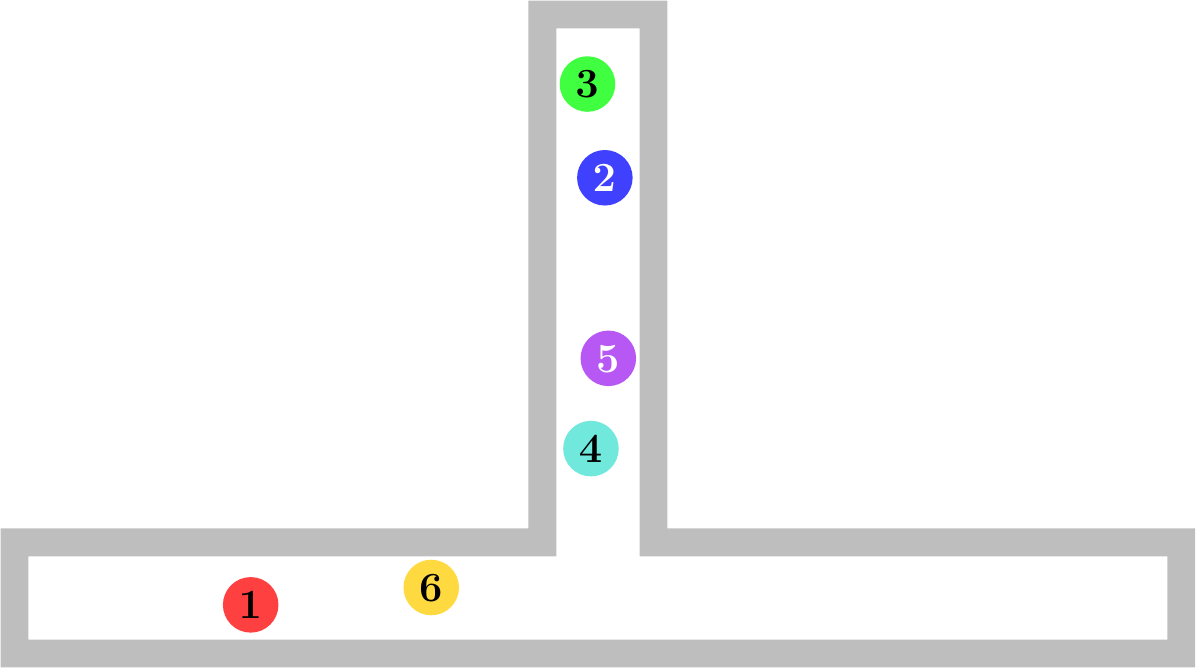}
		\label{subfig:tunnel_env_perm}
	}
    \caption{\fcap Tunnel scenario. The environment consists of a T-shaped
      free space and requires the robots in one side to exchange
      places with the robots on the other side.  There are 6
      translating disc robots of radius 2 and the width of each arm is
      5, so the robots cannot exchange places within an arm without
      leaving it. (a)~Start configuration. The red, blue and
      green robots lie on the left arm, and the yellow, purple and
      cyan robots lie on the right arm. 
      In the goal configuration 
      the red, blue and green robots lie on the right arm
      and the yellow, purple and cyan robots lie on the left arm.
      More specifically,
      the red robot exchanges places with the cyan robot,
      the blue robot with the purple robot
      and the green robot with the yellow robot.
	  (b)~A configuration for which the permutation in $A_U$ is $\left(3,2,5,4\right)$, 
      in $A_R$ is $\left(\;\right)$ and in $A_L$ is $\left(1,6\right)$.
      The corresponding EC is
      denoted by $\left[\left(3,2,5,4\right), \left(\; \right), \left(1,6\right)\right]$.
      }
    \label{fig:tunnel_env}
\end{figure}

\noindent 
%\minisection{\tunnelcl}
As an example of
$\XX_{\textup{\tunnelcl}}$ consider the ``Tunnel'' scenario depicted
in \cref{fig:tunnel_env}. The workspace consists of three portions
corresponding to the three ``arms'' of the workspace: upper arm, right
arm and left arm, denoted by $\calA=\left\{A_U,A_R,A_L\right\}$.  In
this substructure we define the ECs to correspond to
the assignment of robots to portions of the tunnel, and to the
specific order of the robots within each portion. The order in the
upper arm $A_U$ is calculated according to decreasing $y$ coordinate,
in the right arm \mbox{$A_R$ according} to decreasing $x$ coordinate,
and in the left \mbox{arm $A_L$} according to increasing $x$ coordinate.
%the order in the right arm $A_R$ is determined according to decreasing $x$ 
%and the order in the right and left arms ($A_R$ and $A_L$) is determined according to decreasing and increasing $x$ coordinate, respectively.
See \cref{subfig:tunnel_env_perm} for an illustration.

Two ECs are neighbors if they correspond to a transition of a single
robot that leaves one arm and enters another. For instance,
$\left[\left(3,4,2\right),\left(5,6,1\right),\left(\;\right)\right]$
and
$\left[\left(3,4,2,1\right),\left(5,6\right),\left(\;\right)\right]$
are neighbors. This condition implicitly induces the equivalence graph
$G_{\XX_{\textup{\tunnelcl}}}$ and the corresponding natural metric
\dk. For instance, for any two configurations $U,V$ which lie in the
ECs
$\left[\left(3,4,2,5,6,1\right),\left(\;\right),\left(\;\right)\right]$
,
$\left[\left(3,4,1,6,5,2\right),\left(\;\right),\left(\;\right)\right]$
, respectively, it follows\footnote{The shortest path over
  $G_{\XX_{\textup{\tunnelcl}}}$ can be obtained in the following
  manner: \begin{enumerate*}[(1)]
  \item $r_1: A_U\rightarrow A_R$ (namely, $r_1$ moves from the upper
    arm to the right arm) \phantom{}
  \item $r_6: A_U\rightarrow A_R$ \phantom{}
  \item $r_5: A_U\rightarrow A_R$ \phantom{}
  \item $r_2: A_U\rightarrow A_L$ \phantom{}
  \item $r_5: A_R\rightarrow A_L$ \phantom{}
  \item $r_6: A_R\rightarrow A_L$ \phantom{}
  \item $r_1: A_R\rightarrow A_U$ \phantom{}
  \item $r_6: A_L\rightarrow A_U$ \phantom{}
  \item $r_5: A_L\rightarrow A_U$ \phantom{}
  \item $r_2: A_L\rightarrow A_U$.
\end{enumerate*}} that $\dk(U,V)=10$.

\begin{figure*}[t]
  \centering
  \includegraphics[width=0.70\textwidth]{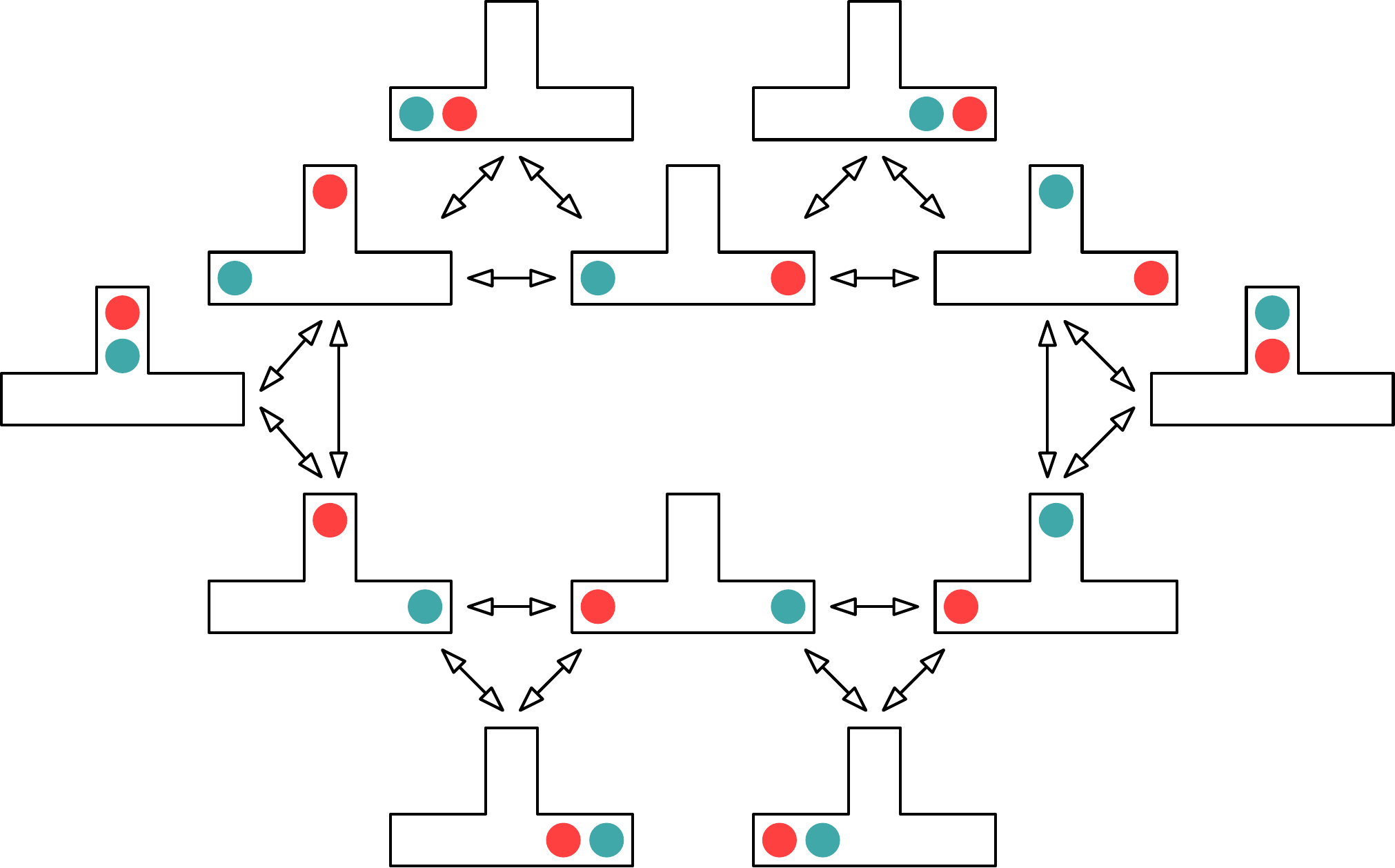}
  \caption{\fcap $G_{\XX_{\textup{\tunnelcl}}}$ for two robots
    ($m=2$). Each vertex of $G_{\XX_{\textup{\tunnelcl}}}$ represents an EC in the joint \Cs $\calX^m$. }
    \label{fig:tunnel_graph}
\end{figure*}

An illustration for the equivalence graph for the case of $m=2$ robots is depicted in \cref{fig:tunnel_graph}.

\subsection{\chamberscl}
\begin{figure}
\ifx\arxiv\undefined
\else
\fi
    \centering
	\subfloat[\fcap]
	{
		\includegraphics[width=0.22\textwidth]{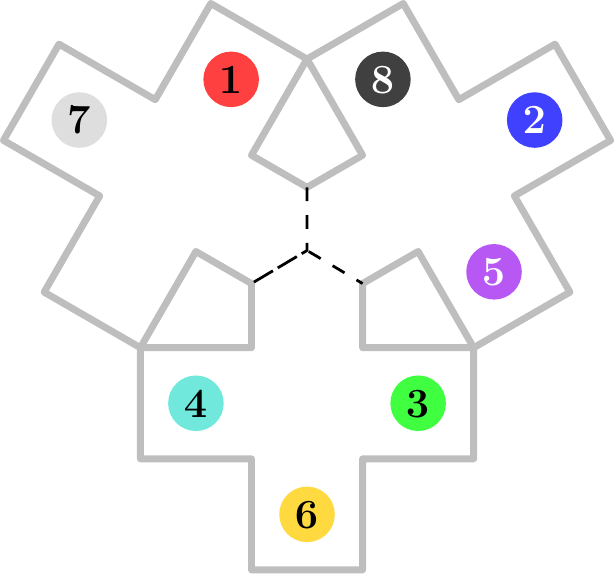}
		\label{subfig:chambers_env_start}
	}
	\quad
	\subfloat[\fcap]
	{
		\includegraphics[width=0.22\textwidth]{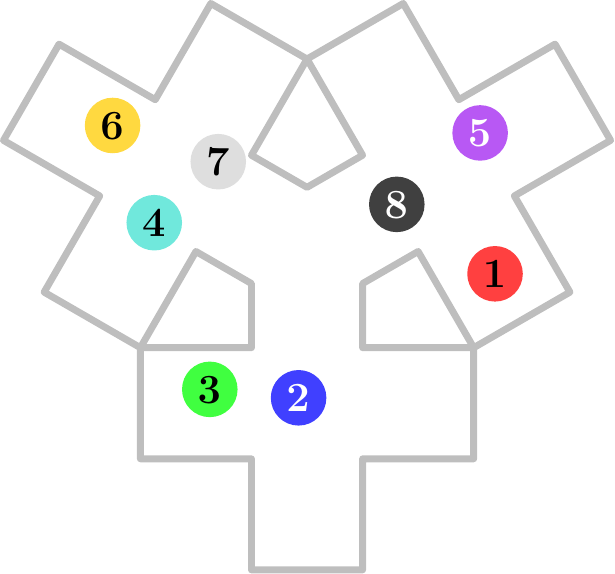}
		\label{subfig:chambers_parted}
	}
	
    \caption{\fcap Chambers scenario. The environment consists of three
      chambers. The structure of each chamber allows the robots to exit
      from the chamber in any order, not necessarily in the order they
      entered to the chamber (as opposed to the arms in the Tunnel scenario).
      (a)~Start configuration. (b)~A configuration that corresponds to the assignment
      $\left[\left\{1,5,8\right\}, \left\{4,6,7\right\}, \left\{2,3\right\}\right]$.
      The natural distance between it and the configuration in
      \cref{subfig:chambers_env_start} is 4.}      
    \label{fig:chambers_env}
\end{figure}

\noindent 
%\minisection{\chamberscl}
As an example of $\XX_\textup{\chamberscl}$ we consider the
``Chambers'' scenario depicted in \cref{fig:chambers_env}.
Each EC is associated with a partitioning of the robots
to the chambers.  Each robot is mapped to the chamber that has the
largest intersection with the robot and we choose a chamber at random
in case that there is a tie. See \cref{subfig:chambers_parted}.  Two
ECs are neighbors if exactly one robot changes its
mapped chamber.  Unlike the previous substructure, here the exact
order of the robots inside one chamber does not matter.

\subsection{\puzzlecl} 
\begin{figure}
	
	\centering
%	\begin{subfigure}[b]{\textwidth}
%        \centering
%        \includegraphics[width=0.12\textwidth]{figures_puzzle_start.pdf}
%        \label{subfig:puzzle_env}
%        \caption{Lorem ipsum}
%    \end{subfigure}
%    
%    \begin{subfigure}[b]{\textwidth}
%        \centering
%        \includegraphics[width=0.12\textwidth]{figures_puzzle_perm.pdf}
%        \label{subfig:puzzle_othr}
%        \caption{Lorem ipsum}
%    \end{subfigure}

	\subfloat[\fcap]
	{
		\includegraphics[width=0.14\textwidth]{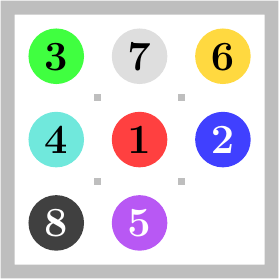}
		\label{subfig:puzzle_env}
	}
	\qquad
	\subfloat[\fcap]
	{
		\includegraphics[width=0.14\textwidth]{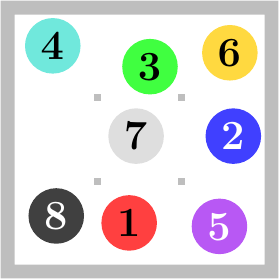}
		\label{subfig:puzzle_othr}
	}
	
    \caption{\fcap 8-Puzzle scenario. The environment can be naturally
      partitioned into nine cells that form a 3$\times$3 grid.  A robot
      can translate only between adjacent cells. (a)~Start
      configuration. The goal is to arrange the robots in the order
      $r_1,\ldots,r_m$, i.e., $r_1$ is situated in the cell in the top
      left corner, and so on. (b)~A configuration that corresponds to
      the assignment
      $\left[\left\{4\right\}, \left\{3\right\}, \left\{6\right\},
        \left\{\;\right\}, \left\{7\right\}, \left\{2\right\},
        \left\{8\right\}, \left\{1\right\}, \left\{5\right\}\right]$.
      The natural distance between it and the configuration in
      \cref{subfig:puzzle_env} is 5 since there is
      a discrete motion
      % with 5 steps
      mimicking five transfer steps each of a single pebble from one cell to another in an 8-puzzle,
      that transforms one configuration to the other (the
      motion involves the purple, red, grey, green and cyan robots).}
    \label{fig:puzzle_env}
\end{figure}

\noindent
%\minisection{\puzzlecl}
The ``8-Puzzle'' scenario, which is a
geometric variation of the classic 15-Puzzle~\citep{10.2307/2589612},
is used as an example for $\XX_\textup{\puzzlecl}$. The problem is
depicted in \cref{fig:puzzle_env}.  Unlike the discrete version of the
puzzle, where each robot can occupy only one of nine possible places,
in the geometric generalization the robots can lie in any
collision-free configuration.

Each EC of $\XX_\textup{\puzzlecl}$ is associated with
an assignment of robots to the nine cells. The cell corresponding to
each robot is the one that has the largest intersection with the
robot, with the restriction that at most one robot is assigned to a
single cell, and we choose a cell at random in case that there is a
tie. An example for a configuration along with its correspondent
assignment is described in \cref{subfig:puzzle_othr}. Two ECs of $\XX_\textup{\puzzlecl}$ are neighbors if exactly one robot
changes its cell assignment.

%"a discrete motion with 5 steps" not well defined; here is an attempt at improving it:
%"mimicking five transfer steps each of a single pebble from one cell to another in an 8-puzzle"

%%% Local Variables:
%%% mode: latex
%%% TeX-master: "../paper"
%%% End:

\section{Analysis of Metrics}
\label{sec:analysis} \noindent
In this section we introduce two novel tools for analyzing metrics,
which rely on the concept of canonical substructures, described
in~\cref{sec:sub}.  The following tools assess the quality of a given
metric $\dds$ by quantifying its similarity to the natural metric
$\dk$, and by counting the number of explored ECs by a
planner that is paired with $\dds$.

In addition to the tools described in this section, we have a
visualization tool that automatically generates an animation for the
expanded tree produced by the RRT-style planner. Some properties of the metrics can be inferred by
perusing the animations. This tool was essential in the screening
phase and guiding our choice of metrics.
Links to example videos can be found in
\ifx\arxiv\undefined
the extended version of this paper.
\else
Appendix~\ref{supp:visualization}.
\fi

\subsection{Distributions separation}\label{subsec:dists} \noindent
The following technique requires as an input, after fixing a specific
canonical substructure $\XX$, a set of $\ell$ randomly sampled joint
configurations $\CC=\{C_1,\ldots,C_\ell\}$ from $\XX$. Each such
sample is then classified according to its EC in $\XX$.

Our working hypothesis is that a good metric should faithfully reflect the natural distance, and in
the rest of the subsection we spell out what it means to have this property.

When incorporating the metric into a sampling-based planner, the role
of the metric is to compare distances between different pairs of
sampled configurations.  Given two pairs of configurations
$\left(U_1,V_1\right)$ and $\left(U_2,V_2\right)$, the planner favors
to check the continuous motion between the first pair in case the
distance between $U_1,V_1$ is smaller than the distance between
$U_2,V_2$ (Note that in the case of an RRT-style planner, the
compared pairs always satisfy $U_1=U_2$.).
%Reflecting
How much a metric reflects the natural distance can be measured by how 
well the relation between distances of different pairs of configurations 
is preserved when compared to the natural distance.
Preserving the natural distance can be measured by
$\Gamma_\dds$:
\begin{align*}
\Gamma_\dd = 
\underset{U_1,U_2,V_1,V_2\in\XX}{\Pr}\Bigl[
&\dd\left(U_1,V_1\right) < \dd\left(U_2,V_2\right)
\Bigm\vert\;
\Bigr. \\ \Bigl.
&\dk\left(U_1,V_1\right) < \dk\left(U_2,V_2\right)
\Bigr].
\end{align*}

In one extreme case, if we use the natural distance as $\dds$ we have
$\Gamma_\dds=1$.  In the other extreme case, if a metric $\dds$ has no
correlation with the natural distance we have $\Gamma_\dds=0.5$.  We
are interested in a metric that gives a large value of
$\Gamma_\dds$.

In the rest of the subsection we formalize the discussion above and
explain how to calculate and compare $\Gamma_\dds$ between different
metrics.  For every possible (discrete) value of the natural distance
$\alpha \in\Ima \dk$ we compute the set ${\calD_\dd}^\alpha$ of metric
distances given that the natural distance is $\alpha$:
\begin{equation*} 
  {\calD_\dd}^\alpha = \left\{\dd\left(U,V\right) \mid
    U,V\in\mathbb{C}, \dk\left(U,V\right) = \alpha \right\}.
\end{equation*}

With a slight abuse of notation, we treat ${\calD_\dd}^\alpha$ as a
distribution over pairs of configurations from $\XX$. Here we use the
fact that $\CC$ captures the structure of $\XX$. 
Furthermore, we define $
\calD_\dd = \left\{ {\calD_\dd}^\alpha \mid \alpha \in \Ima \dk \right\}
$.
Consequently, $\Gamma_\dds$ can be represented as
\begin{equation*}
  \Gamma_\dds=\Pr\Bigl[
  \alpha_0 < \beta_0 
  \;\Bigm\vert\; 
  \alpha_0 \sim {\calD_\dd}^\alpha, 
  \beta_0 \sim {\calD_\dd}^\beta, 
  \alpha < \beta
  \Bigr], 
\end{equation*}
where the notation $\alpha_0 \sim {\calD_\dd}^\alpha$ indicates that
$\alpha_0$ is sampled from the distribution ${\calD_\dd}^\alpha$.

Sampling-based planners usually attempt to connect nearby
configurations. Thus, it is more important to identify close
configurations than remote ones.
Pairs of far-away configurations (with respect to the natural distance) are practically ignored by a sampling-based planner that uses a reasonable metric $\dds$.
We restrict $\Gamma_\dds$ to natural
distances of at most a threshold parameter $\tau$, using the following
definition\footnote{We require that
  $\alpha\leq \tau$, and not $\beta$, since we only care that pairs of
  configurations with small value of $\textup{d}_{\calK}$ will remain
  so with respect to $\textup{d}$. A similar correlation is not
  assumed between large distances.} of ${\Gamma_\dd}^\tau$:
\begin{equation*}
{\Gamma_\dd}^\tau=\Pr\Bigl[
\alpha_0 < \beta_0 
\;\Bigm\vert\; 
\alpha_0 \sim {\calD_\dd}^\alpha, 
\beta_0 \sim {\calD_\dd}^\beta, 
\alpha < \beta,
\alpha \leq \tau
\Bigr].
\end{equation*}
We expect that a metric $\dds_1$ will be more effective than a metric
$\dds_2$ if ${\Gamma_{\dds_1\!}}^\tau > {\Gamma_{\dds_2\!}}^\tau$.

  Note that the value of~$\tau$ depends on the specific
  setting. Here we present general guidelines for choosing~$\tau$. An exact method is
  left for future work. Recall that in each iteration of an RRT-style planner a
  configuration~$V$ is sampled at random, and its nearest-neighbor~$U$ (from the
  currently growing tree) is picked. A proper value for~$\tau$
  satisfies~$\dk\left(U,V\right)\leq \tau$ with high probability for a
  typical RRT-tree size.
  Pairs of configurations for which the natural distance is larger than~$\tau$ are
  practically ignored by a sampling-based planner, and should not be taken into account
  in the calculation of~$\Gamma_\dds$.
%  Using a larger value for~$\tau$ will incorporate
%  irrelevant data into the calculation of~$\Gamma_\dds$.}

\subsection{Explored equivalence classes}\label{subsec:explored} \noindent
RRT-style planners, as the one used and described later on in
\cref{sec:experiments}, explore the \Cs from a starting
configuration. A desirable property of such planners is to reach
various regions of interest in the \Cs.  In our setting, we measure
the quality of exploration by the number of different ECs reached,
where a larger number of explored ECs means that the planner explores
the \Cs more exhaustively.  Since the planner cannot foresee which
parts of the \Cs can lead to a solution, we expect that an effective
metric will result in a larger number of explored ECs when compared to
an ineffective one.

We propose the following experiment to assess $\dds$ with respect to
the quality of exploration. A single-tree RRT-style planner is used to
build a tree with $N$ vertices.  The set of explored configurations is
denoted by $\calU_\dds$.  For each configuration $U\in\calU_\dds$ we
identify its representative EC denoted by $\EC\left(U\right)$.  
We count the
number of distinct explored ECs, i.e., the number of
distinct ECs in the set 
$\left\{\EC\left(U\right) \;\vert\; U \in
    \calU_\dds\right\}$,
and denote it by $\left|\calU_\dds / \EC\right|$. 
We anticipate that a metric $\dds_1$ will be more effective than a
metric $\dds_2$ if
$\left|\calU_{\dds_1} / \EC\right| > \left|\calU_{\dds_2} /
  \EC\right|$.

%%% Local Variables:
%%% mode: latex
%%% TeX-master: "../paper"
%%% End:

\section{Experimental Results}
\label{sec:experiments} \noindent In this section we make use of the
tools developed in \cref{sec:analysis} to analyze the properties of
the metrics in the scenarios described in \cref{sec:sub}. Then we compare the effectiveness of
the metrics as used by dRRT~\citep{DBLP:journals/ijrr/SoloveySH16} to
solve instances of MRMP.  As mentioned in \cref{sec:screening}, dRRT
is an extension of RRT, which allows it to cope with a greater number of
robots and more complex scenarios.  Later on we show the effectiveness
of the planner incorporated with different metrics in a general
environment that consists of several substructures.

On the implementation side, our testing environment is
implemented in C++ and relies on \emph{Open Motion Planning
  Library (\ompl)}~\citep{DBLP:journals/ram/SucanMK12}. While we
are not concerned with running times in this work, we mention
that all the metrics defined in \cref{sec:metrics} can be implemented with running time linear in the number of robots.
Refer to 
\ifx\arxiv\undefined
the extended version of this paper
\else
Appendix~\ref{supp:calculation}
\fi
for full description of the implementation.

% \subsection{Implementation details} \noindent
% All the experiments 
% %during the study 
% were performed using a
% cluster of 40 single-core virtual machines running over Google Compute
% Engine~\citep{google:cloud}.  
% %Each virtual CPU is implemented as a single hardware hyper-thread on a 2.3 GHz Intel Xeon E5 v3. 

% \reviewer{The comment about running a sequential algorithm on a cluster is not relevant
% when no times are reported.} \aviel{I think we should suppress
% any mention for running times since it can open a window to our
% unsolved problem of NN-search in high dimensional spaces.}
% \kiril{I agree. I've already done that in places that I found
%   relevant. Make sure that it doesn't appear elsewhere.}

\subsection{Analyzing properties of the metrics} \noindent
We show and analyze the results of the experiments described in \cref{sec:analysis} using the scenarios described in \cref{sec:sub}.

For each scenario, we show results for the value of~${\Gamma_\dd}^\tau$ defined in \cref{subsec:dists}.  Then, we count
the number of distinct explored ECs, as suggested in
\cref{subsec:explored}.  In order to do so, we use a dRRT-tree with
10,000 vertices rooted at the start configuration (see
\cref{subfig:tunnel_env_start,subfig:puzzle_env,subfig:chambers_env_start})
.  Finally, we show the effectiveness of an entire planning algorithm
that uses each of the metrics and show how it correlates with the
results of the analysis tools.  
We measure the
effectiveness of the planner by inspecting both~\begin{enumerate*}[(i)]
\item the number of explored vertices when a solution is found---the lower the number, the
more effective we consider the metric to be;
\item the success rate of the planner.
\end{enumerate*}
We mention that the success rate of the local planner 
(and not the motion planner) 
is similar among all the metrics, and therefore we do not report it.\footnote{As discussed in \cref{sec:screening}, metrics that induce high success rate of the local planner are not necessarily effective for planning.}
%can be used as
%an alternative measure. However, from our experience it leads to
%results similar when using the number of explored vertices. 
%Therefore,
%the success rate of local planning is not reported. 
We do not measure running times since we are interested only in
the analytic effectiveness of each metric.

\begin{figure*}
    \centering
	\subfloat[\fcap \tunnelcl\xspace Substructure]
	{
		\includegraphics[width=0.29\textwidth]{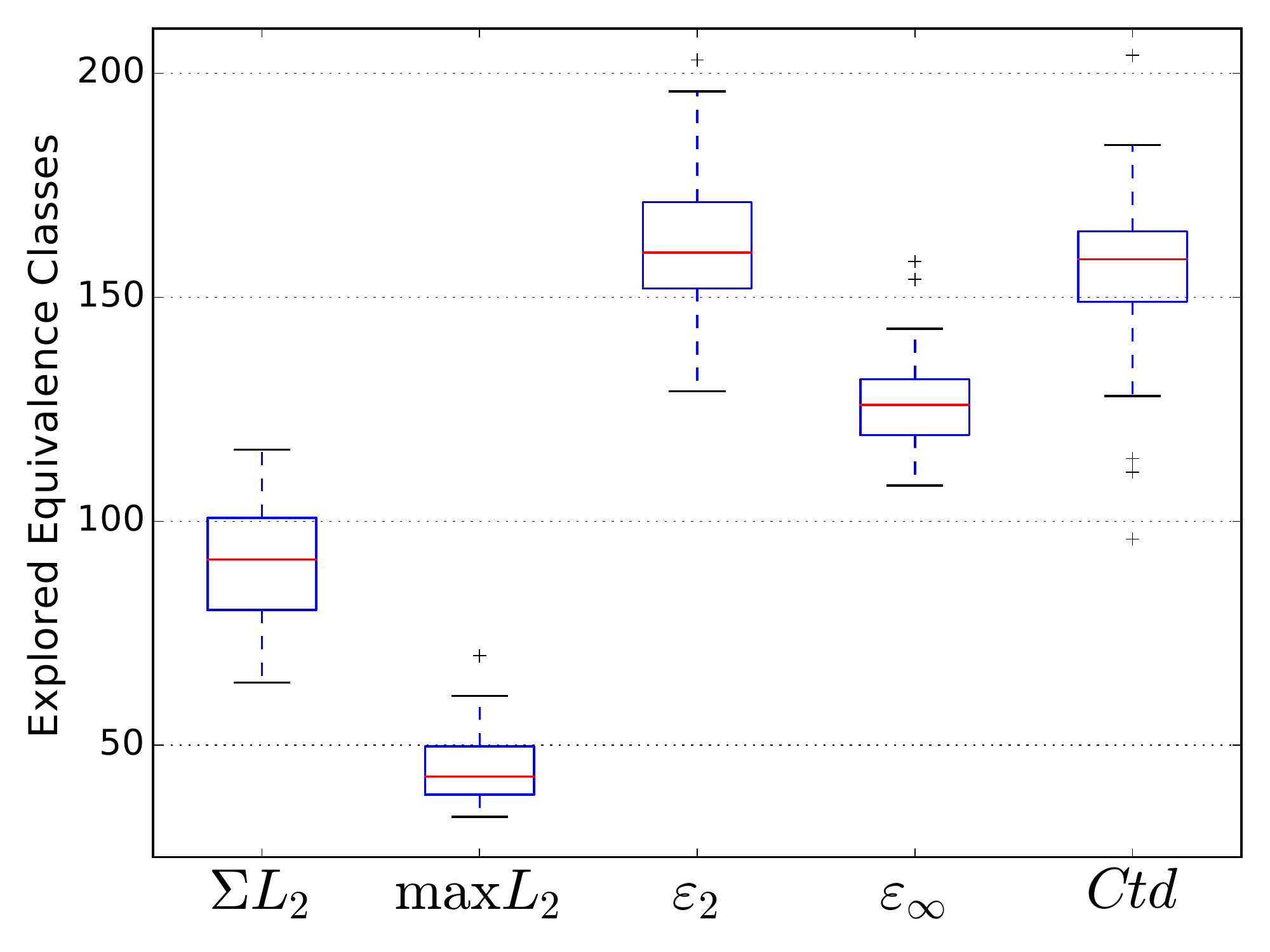}
		\label{subfig:explored_tunnel}
	}
	\subfloat[\fcap \chamberscl\xspace Substructure]
	{
		\includegraphics[width=0.29\textwidth]{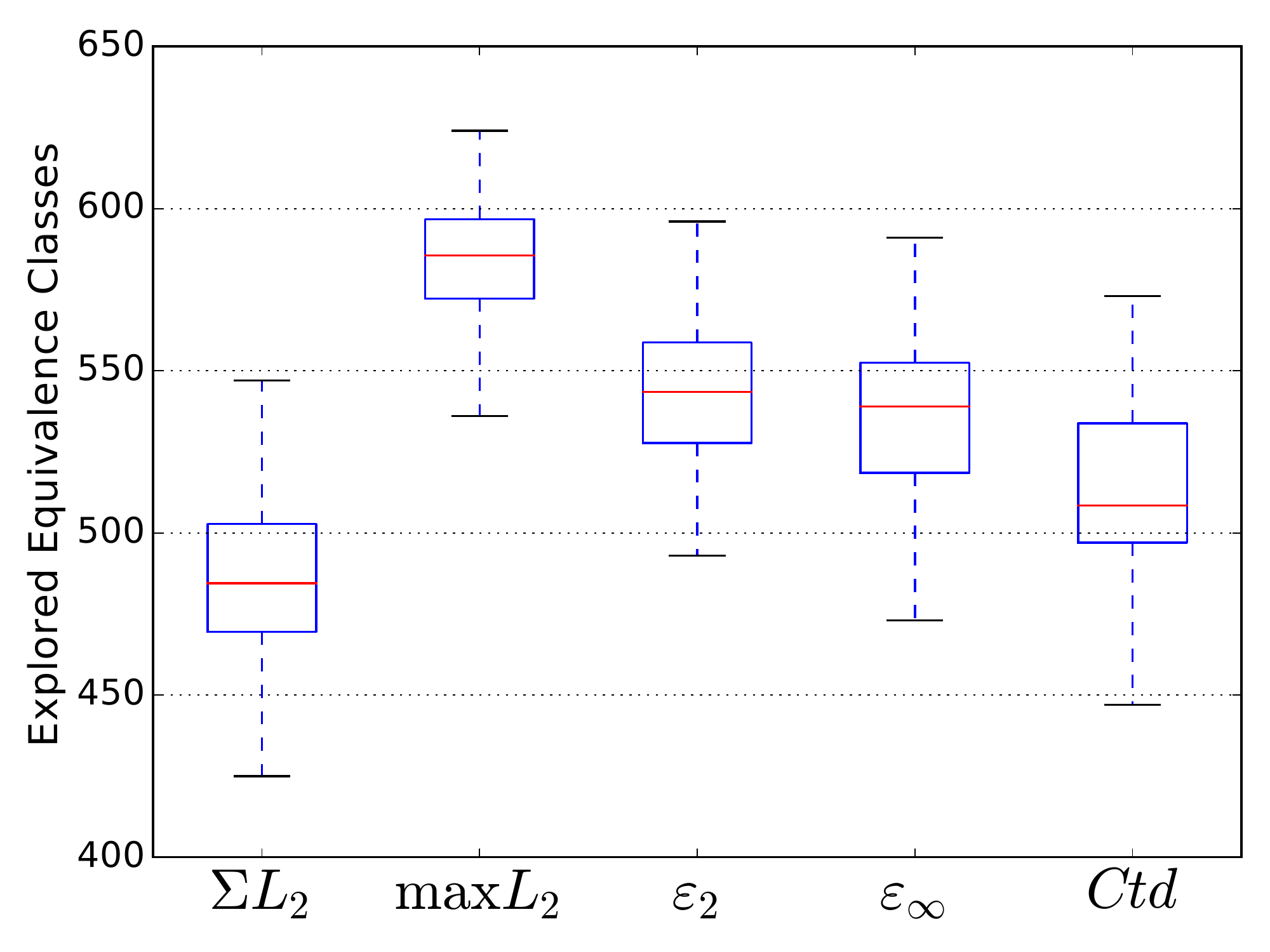}
		\label{subfig:explored_chambers}
	}
	\subfloat[\fcap \puzzlecl\xspace Substructure]
	{
		\includegraphics[width=0.29\textwidth]{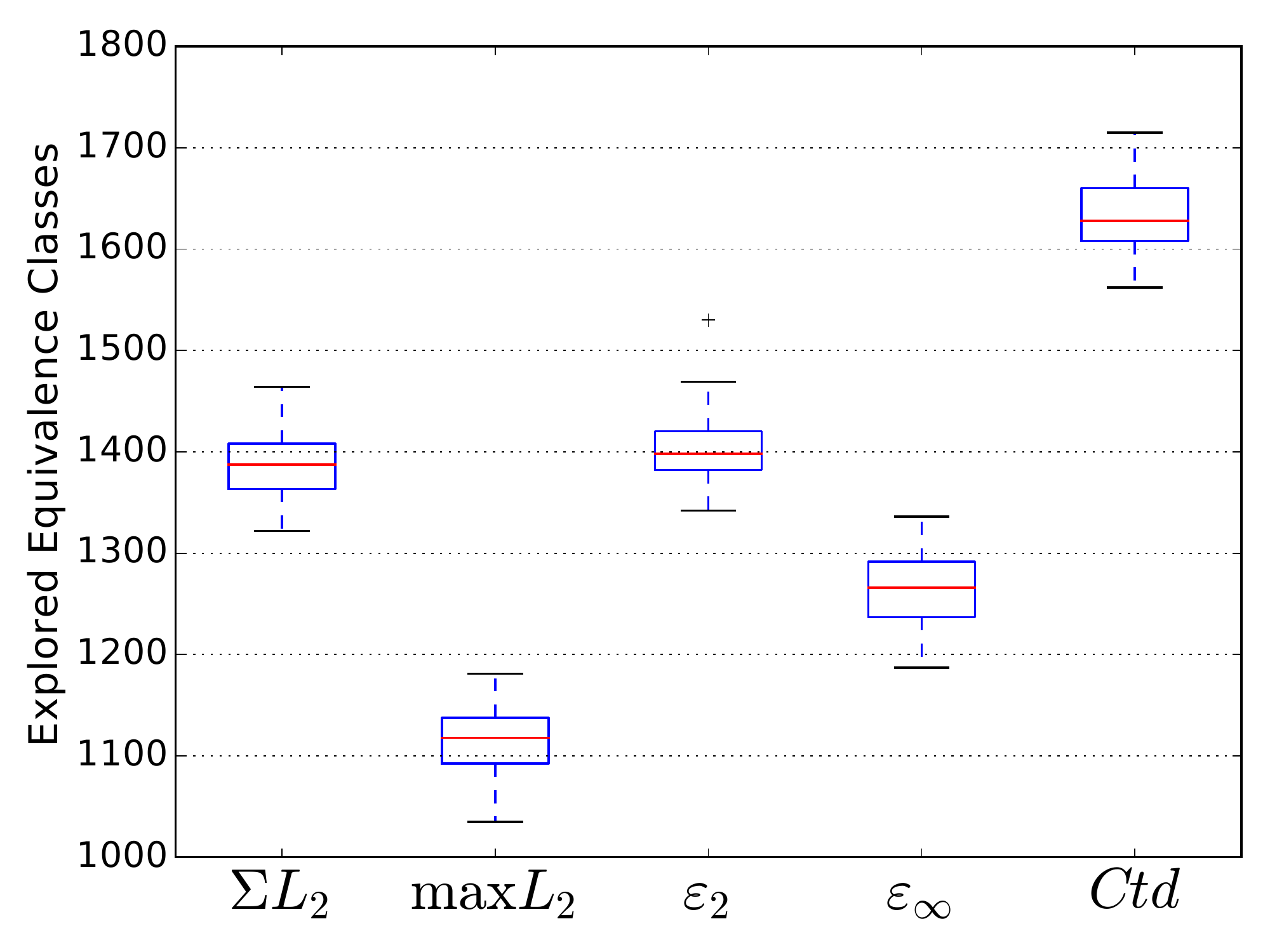}
		\label{subfig:explored_puzzle}
	}
	
    \caption{\fcap Different explored equivalence classes experiment. A dRRT tree is expanded until it contains 10,000 vertices. For each vertex in the tree we find its representative EC, and count the number of different ECs (denoted by $\left|\calU_\dds/\EC\right|$). Higher value means that we expect the metric to be more effective.
     %Then, $\left|\calU_\dds/\EC\right|$ is calculated for each metric. 
     The experiment is repeated 50 times for each metric. The figure depicts the values of $\left|\calU_\dds/\EC\right|$ for each metric.}
    \label{fig:experiments_explored}
\end{figure*}

\begin{figure*}
    \centering   
	\subfloat[\fcap \tunnelcl\xspace Substructure]
	{
		\includegraphics[width=0.29\textwidth]{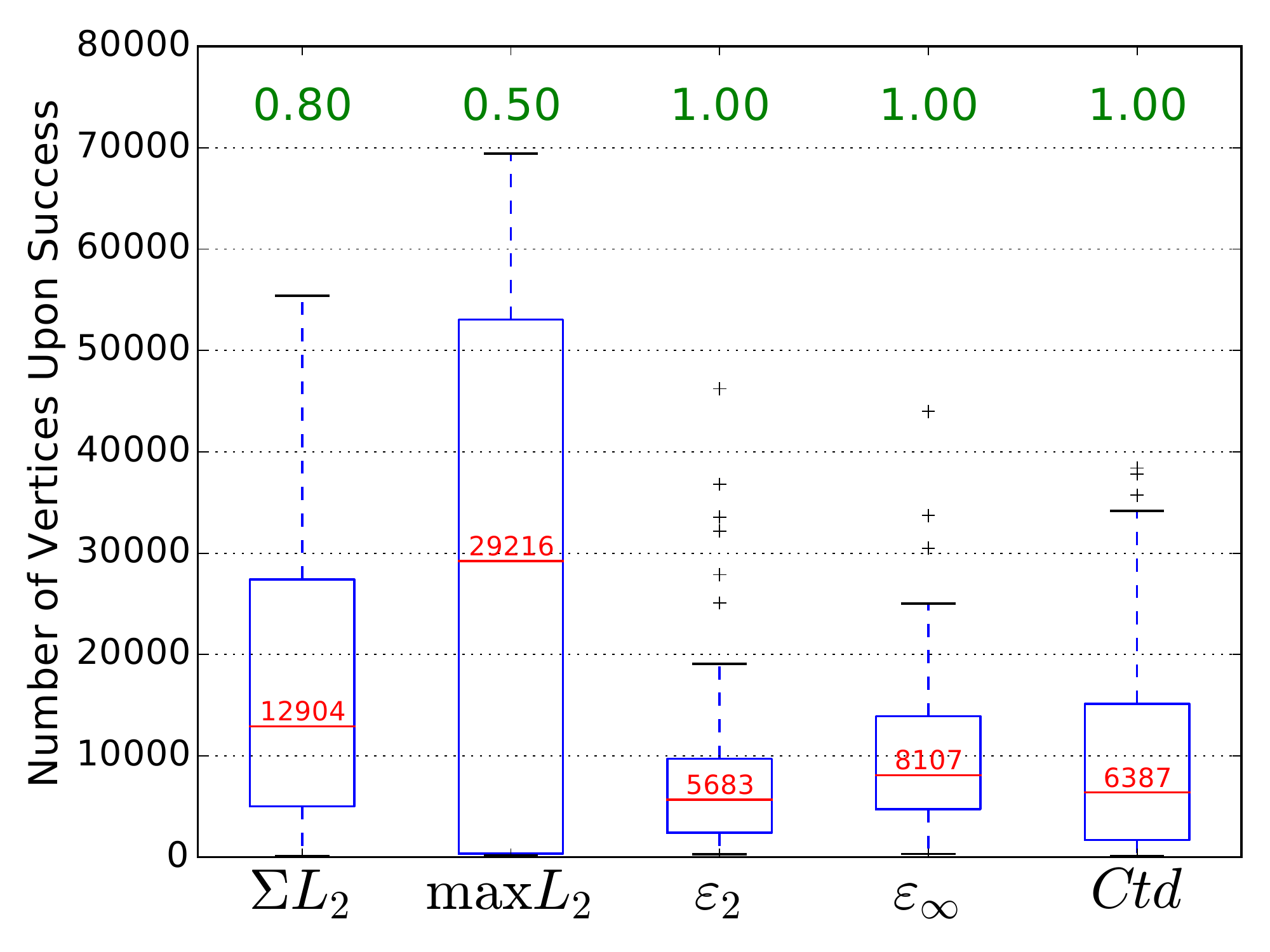}
		\label{subfig:drrt_tunnel}
	}
	\subfloat[\fcap \chamberscl\xspace Substructure]
	{
		\includegraphics[width=0.29\textwidth]{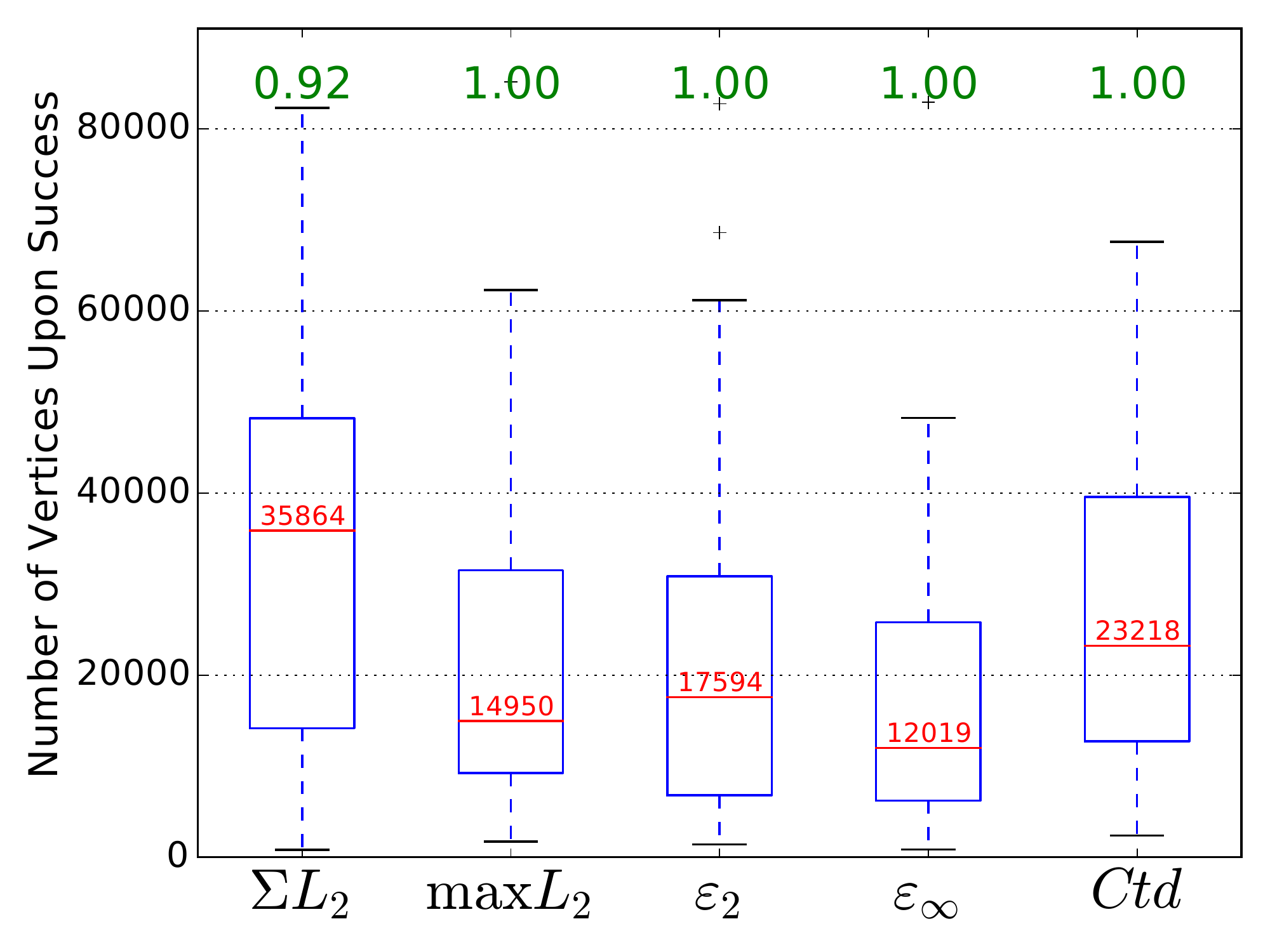}
		\label{subfig:drrt_chambers}
	}
	\subfloat[\fcap \puzzlecl\xspace Substructure]
	{
		\includegraphics[width=0.29\textwidth]{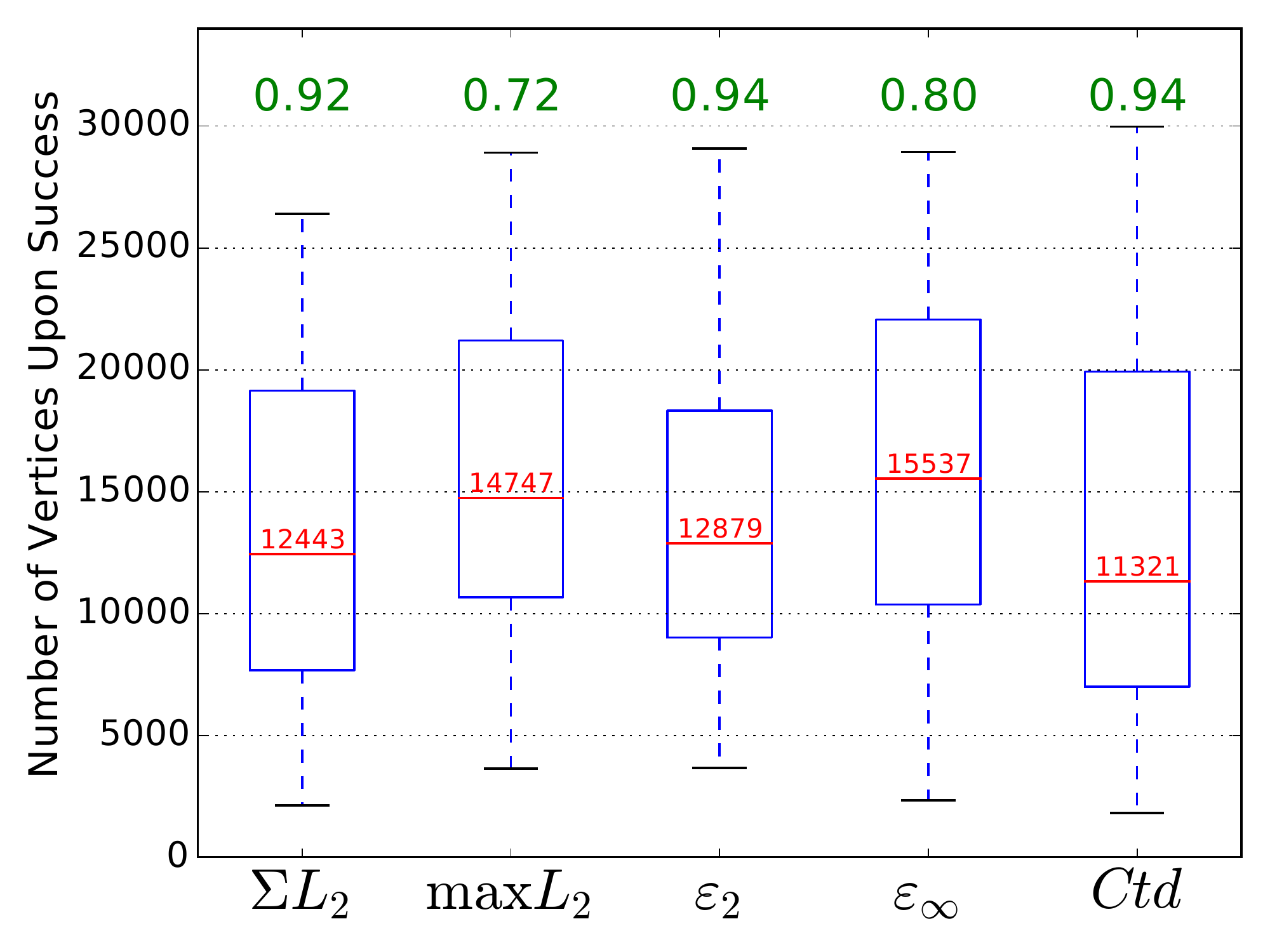}
		\label{subfig:drrt_puzzle}
	}
	
    \caption{\fcap Number of expanded vertices when a solution is found. 
    The experiment is repeated 50 times per metric. 
    The planner success rate is depicted in the green labels on top of each boxplot. 
    The red labels are the median value.
    Effectiveness is expressed by high success rate and low number of vertices.}
    \label{fig:experiments_drrt}
\end{figure*}

Next, for each typical scenario we describe (i)~the results of the
distributions-separation predicates, (ii)~the results of the ECs
exploration, and finally (iii)~the actual behavior of the planner and
its relation to the predictions. These are also summarized in
\cref{tab:results_dists}, \cref{fig:experiments_explored} and
\cref{fig:experiments_drrt}, respectively.

\vspace{5pt}

\begin{figure}
    \centering
	\subfloat[\fcap $\calD_{\SLT}$]
	{
		\includegraphics[width=0.23\textwidth]{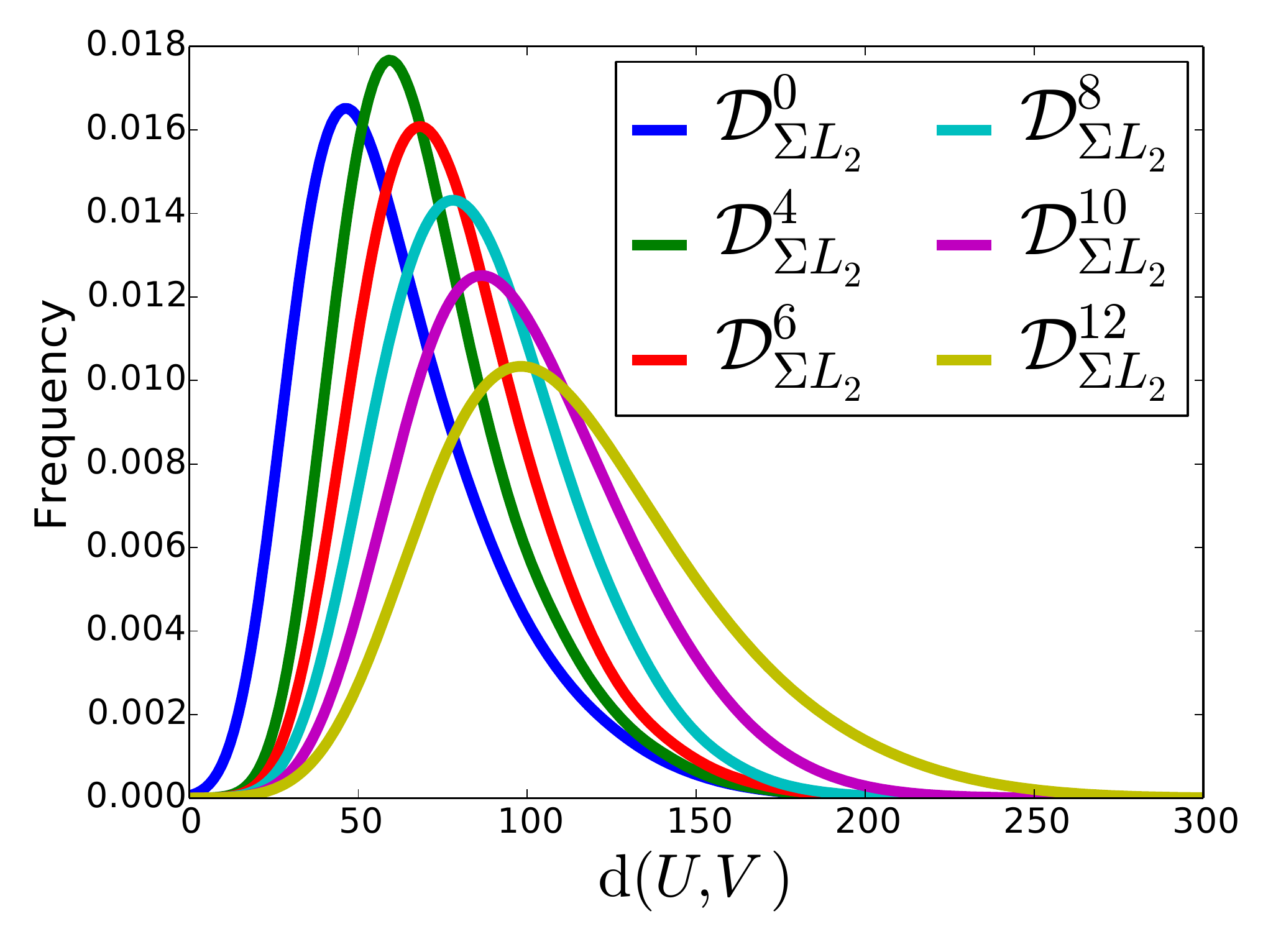}
	}
	\subfloat[\fcap $\calD_{\epst}$]
	{
		\includegraphics[width=0.23\textwidth]{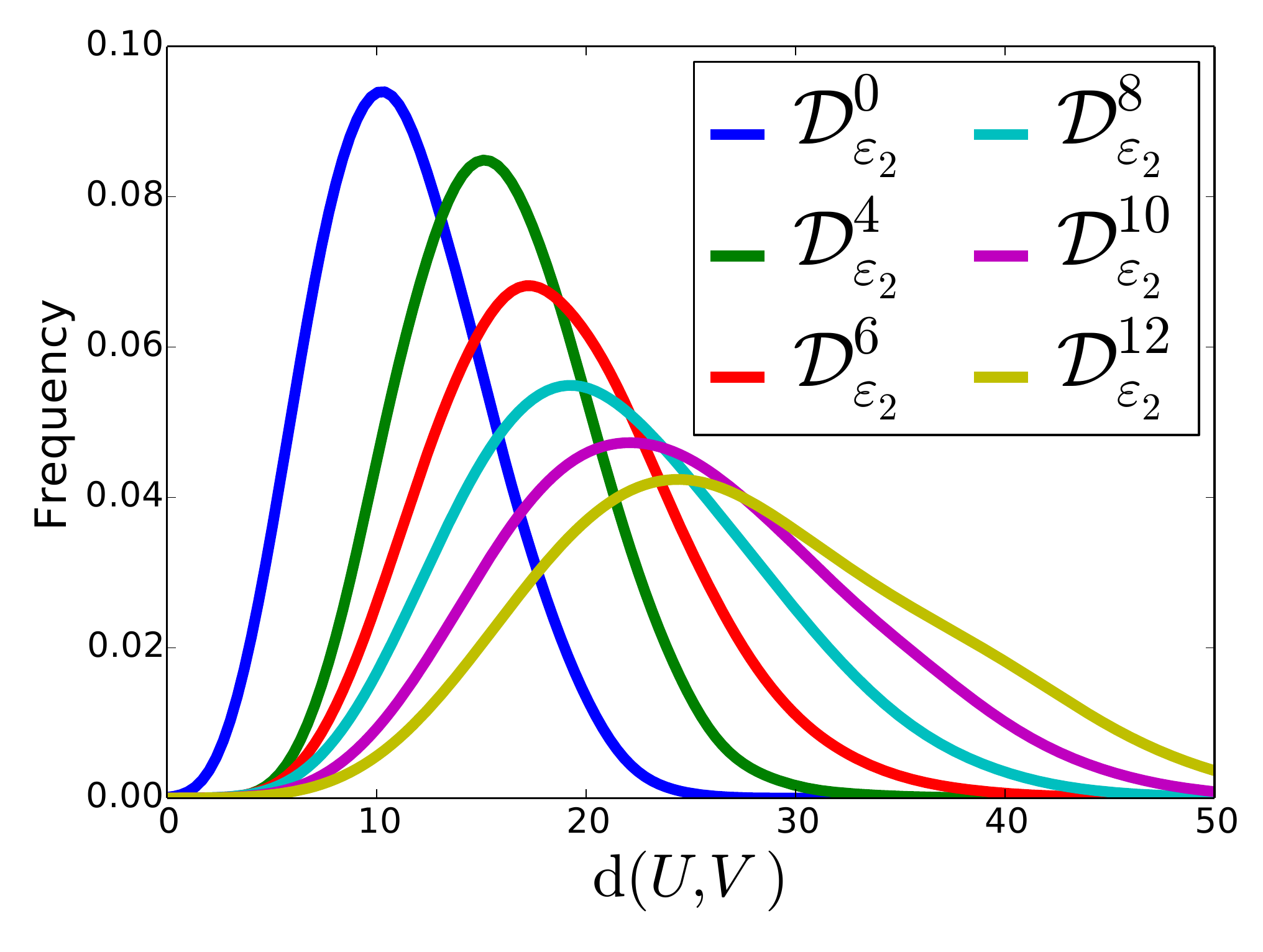}
	}
	
    \caption{\fcap Distributions from $\calD_\dds$ for \SLT and \epst
      metrics in the Tunnel scenario. Better reflection of the natural
      distance is expressed by higher level of separability between
      the distributions.}
    \label{fig:tunnel_natural_graphic}
\end{figure}

\minisection{\tunnelcl\xspace substructure}
\cref{fig:tunnel_natural_graphic} shows subsets of the sets of
distributions~$\calD_{\epst}$ and~$\calD_{\SLT}$ for the Tunnel
scenario: observe that the distributions in~$\calD_{\epst}$ are better
separated than the distributions in~$\calD_{\SLT}$.  This separation
is expressed by the dissimilarities between the different
distributions. For example, the common area bounded by the blue and
green distributions (representing~${\calD_\dd}^0$ and~${\calD_\dd}^4$
respectively) is smaller for \epst when compared to \SLT. This is also
the case for the green and red distributions (representing~${\calD_\dd}^4$ and~${\calD_\dd}^6$ respectively). The value~${\Gamma_\dd}^\tau$ quantifies the distribution separation.  For this
scenario we set~$\tau=4$. The values of~${\Gamma_\dd}^\tau$ are given
in \cref{tab:results_dists}. The values for \epst, \epsinf and \centr
are similar to each other, and are larger than the values for \SLT and
\Linf.

\begin{table}
  \begin{center}
    \resizebox{0.75\columnwidth}{!}{
      \begin{tabular}{| c | c | c | c | c | c | c | c |}
		\toprule
		\textbf{Scenario} & $\boldsymbol{\tau}$ & \multicolumn{5}{c|}{\textbf{Metric ($\boldsymbol{\dds}$)}} \\
		\midrule
		
		\multicolumn{2}{|c|}{} & \SLT & \Linf & \epst & \epsinf & \centr \\
		\midrule
		
		Tunnel & 4 & 0.810 & 0.843 & 0.904 & 0.904 & 0.907 \\
		\midrule
		
		Chambers & 1 & 0.858 & 0.983 & 0.971 & 0.962 & 0.938 \\
		\midrule
		
		8-Puzzle & 7 & 0.953 & 0.938 & 0.951 & 0.921 & 0.971 \\
		\bottomrule
      \end{tabular}
    }
  \end{center}
  \caption{\tcap The value of ${\Gamma_\dd}^\tau$ for different metrics in different scenarios.
    Each entry in the table is the value of ${\Gamma_\dd}^\tau$ for the corresponding $\dds$, $\tau$ and scenario.
    Larger values mean higher distributions separation, and in turn
    better effectiveness is expected.
    The value of~$\tau$ is set according to the guidelines described in \cref{subsec:dists}}. \label{tab:results_dists}
\end{table}

The number of distinct explored ECs is shown in
\cref{subfig:explored_tunnel}: observe that \centr and
$\eps$-congruence-type metrics show better results when compared to
the standard metrics. In addition, we expect that \epst and \centr
will be more effective than \epsinf. Furthermore, \SLT shows better
results than \Linf.

As described in \cref{subfig:drrt_tunnel}, the effectiveness of the
metrics correlates with the analysis of \cref{sec:analysis}.  As
expected, \epst, \epsinf and \centr are more effective than \SLT and
\Linf.

%%% Local Variables:
%%% mode: latex
%%% TeX-master: "../paper"
%%% End:

%%% Local Variables:
%%% mode: latex
%%% TeX-master: "../paper"
%%% End:

\vspace{5pt}

\minisection{\chamberscl\xspace substructure}
For the distributions separation we use $\tau=1$. The values of ${\Gamma_\dd}^\tau$ are given in \cref{tab:results_dists}.
\Linf has the largest value, then come \epst, \epsinf and \centr, while \SLT is far behind.

\cref{subfig:explored_chambers} shows the number of distinct explored
ECs. \Linf shows the best results, \epst and \epsinf
have comparable results, which are better than \centr, and \SLT yields
the poorest results.

For this scenario, by looking at the results of the experiments described in \cref{sec:analysis}, one can foresee that \Linf, \epsinf and \epst will be more effective than \centr, which in turn, will be more effective than \SLT.
This is indeed the case when measuring the effectiveness of the planner, as can be seen in \cref{subfig:drrt_chambers}.

%%% Local Variables:
%%% mode: latex
%%% TeX-master: "../paper"
%%% End:

\vspace{5pt}

\minisection{\puzzlecl\xspace substructure}
For the calculation of ${\Gamma_\dd}^\tau$ we use $\tau=7$. The values are given in \cref{tab:results_dists}.
The best value is achieved by \centr, then \epst and \SLT have comparable values, then comes \Linf and finally \epsinf with the smallest value.

The number of distinct explored ECs is shown in \cref{subfig:explored_puzzle}.
Here again, the largest number of explored ECs is achieved with \centr, followed by \epst and \SLT. Then \epsinf, and the lowest value is for \Linf.

The effectiveness of the planner incorporated with each metric is
expressed in \cref{subfig:drrt_puzzle}. The results are with
accordance to the analysis: \centr is the most effective metric, \SLT
and \epst have comparable effectiveness, and \epsinf and \Linf are the
less effective metrics.

%%% Local Variables:
%%% mode: plain-tex
%%% TeX-master: "../paper"
%%% End:

%\vspace{5pt}

\subsection{Putting it all together} 
\begin{figure*}
    \centering
	
	\subfloat[\fcap]
	{
		\includegraphics[width=0.23\textwidth]{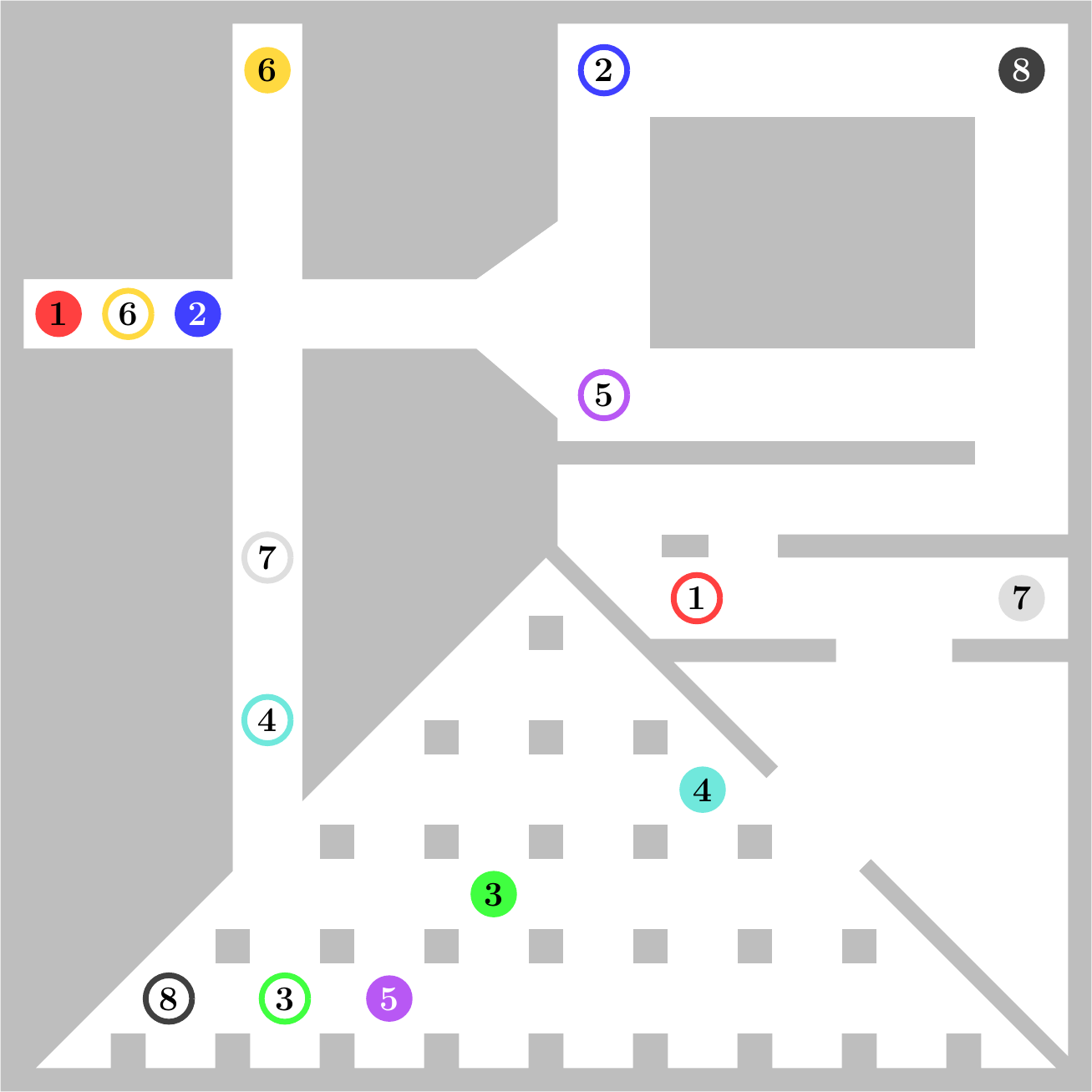}
		\label{subfig:general_start}
	}
	\subfloat[\fcap Effectiveness with 6 robots]
	{
		\includegraphics[width=0.31\textwidth]{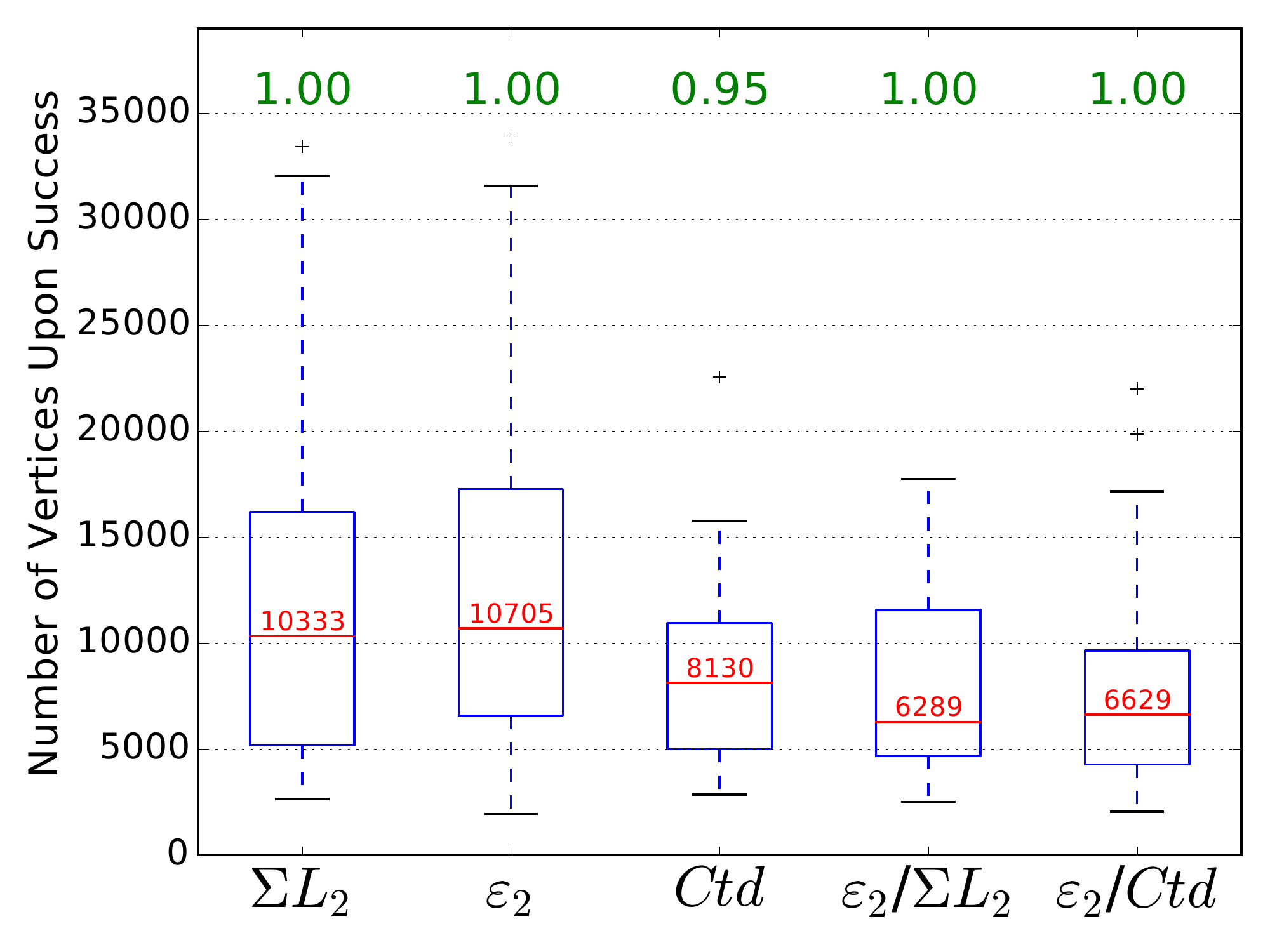}
		\label{subfig:general_6_drrt}
	}
	\subfloat[\fcap Effectiveness with 8 robots]
	{
		\includegraphics[width=0.31\textwidth]{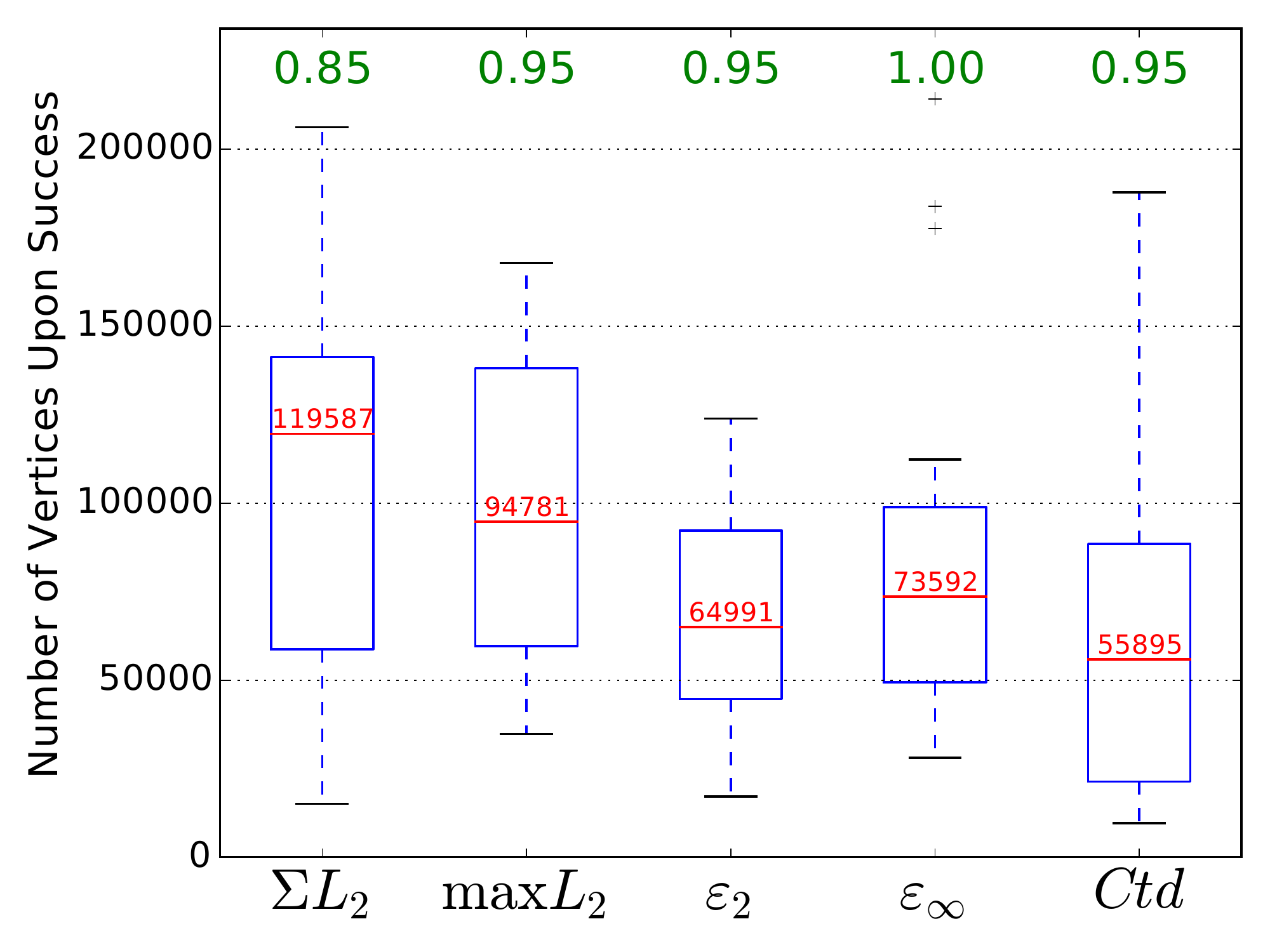}
		\label{subfig:general_8_drrt}
	}
	
    \caption{\fcap A general scenario.  We test the scenario with 8
      robots, and the scenario with 4 or 6 robots which we get by
      eliminating $r_5,\ldots,r_8$ or $r_7,r_8$, respectively.
      (a)~Start and goal configuration, drawn in solid and empty
      discs, respectively.  (b)~Effectiveness of metrics and
      alternation between metrics summarized over 20 runs for the case
      of $6$ robots. As in the previous plots, the green labels
      indicate the success rate. (c)~Effectiveness of each metric
      summarized over 20 runs of the planner for the case of $8$
      robots.}
    \label{fig:general}
\end{figure*}

\noindent The \Cs of a
general MRMP problem may consist of several substructures.  This
is the case for the scenario depicted in
\cref{subfig:general_start}, which contains $m=8$ robots.
\cref{subfig:general_8_drrt} shows the effectiveness of planning
with each metric.  As can be inferred from the results, even in
more general scenarios, the novel metrics are more effective than
the standard ones. In some cases, it may be beneficial to
alternate between several metrics---the planner maintains several
nearest-neighbors data-structures, each for a different
metric. Each time the tree is expanded, a different
data-structure is used in a round-robin fashion.

We have tested the scenario depicted in
\cref{subfig:general_start} with 4, 6 and 8 robots (for 4 and 6
robots we eliminate from the scenario the robots $r_5,\ldots,r_8$
and $r_7,r_8$ respectively).  We used each of the five metrics,
along with all the combinations of two out of the five (total of
$15$) metrics.  For the scenario with $m=4$ robots, the
effectiveness of all the metrics and their alternation was
comparable.  The results for the scenario with $m=6$ robots (see
\cref{subfig:general_6_drrt}) support the claim that it may be
better to alternate between different metrics.  Note the
interesting fact that when alternating between \epst and \SLT or
\centr, better effectiveness is obtained than when using each
metric separately.  For the scenario with $m=8$ robots
(\cref{subfig:general_8_drrt}) the novel metrics are more
effective when compared to the standard ones.  Alternating
between novel and standard metrics does not make the planner more
effective for the case of $8$ robots.  As we move from $4$ robots
(easier) to $8$ robots (considerably harder), the effectiveness
of the metrics becomes more noticeable.

%%% Local Variables:
%%% mode: latex
%%% TeX-master: "../paper"
%%% End:

%%% Local Variables:
%%% mode: latex
%%% TeX-master: "../paper"
%%% End:

\section{Other Multi-Robot Systems}
\label{sec:rotate}\noindent
In this section we show how to extend the metrics that were defined in \cref{sec:metrics} to cope with rotating robots in 2D. We also show experimental results that demonstrate the suitability of the novel metrics for rotating robots.

We mention that all metrics can be extended to 3D settings (with rigid-body motions) straightforwardly. Extensions for other robotics systems, e.g., robots with dynamic constraints, is beyond the scope of this paper and is left for future work.

\subsection{Metrics definition}
Let
$U=\left(u_1,\ldots,u_m\right),V=\left(v_1,\ldots,v_m\right)\in\calX^m$
be two multi-robot configurations. Each single-robot
configuration~$w\in\calX$ can be represented as three coordinates
$x\left(w\right),y\left(w\right),\theta\left(w\right)$ for which
\mbox{$x\left(w\right),y\left(w\right)\in\R$} and
\mbox{$\theta\left(w\right)\in\left[-\pi,\pi\right)$}.  Define:
\begin{align*}
&x_i=x\left(v_i\right)-x\left(u_i\right), \\
&y_i=y\left(v_i\right)-y\left(u_i\right), \\
&\theta_i=
\begin{cases}
\left|\theta\left(v_i\right) - \theta\left(u_i\right)\right|, & \left|\theta\left(v_i\right) - \theta\left(u_i\right)\right| < \pi \\
2\pi - \left|\theta\left(v_i\right) - \theta\left(u_i\right)\right|, & \textup{otherwise}
\end{cases}.
\end{align*}
In addition, we introduce a weight parameter $0\leq s\leq 1$ which determines the weight between the translation and the rotation components.

Having defined $x_i,y_i,\theta_i$, the traditional metrics $\SLT$
and $\Linf$ can be trivially extended:
\begin{align*}
  \Sigma L_2\left(U,V\right) &=
                               \sum_{i=1}^m \left(
                               s \sqrt{x_i^2 - y_i^2} + \left(1-s\right)\theta_i \right), \\
  \max L_2\left(U,V\right) &=
                             \max_{i=1,\ldots,m} \left(
                             s \sqrt{x_i^2 - y_i^2} + \left(1-s\right)\theta_i \right).
\end{align*}

Similarly, we redefine
$\eps$-congruence and $\centr$ by treating the rotational component as an
additional (scaled) coordinate in Euclidean space\footnote{$\mathrm{MiniBall}$ is the radius of
  the smallest enclosing ball, and $\mathrm{MiniCube}$ is the
  radius of the smallest axis-aligned bounding cube of a set of points, defined in the three-dimensional Euclidean space. Refer
  to Appendix~\ref{supp:calculation} for more details.}:
\begin{align*}
\epst\left(U,V\right)=\mathrm{MiniBall}\left(
\left\{\left(s x_i, s y_i,\left(1-s\right)\theta_i\right)\right\}_{i=1}^m
\right), \\
\epsinf\left(U,V\right)=\mathrm{MiniCube}\left(
\left\{\left(s x_i, s y_i,\left(1-s\right)\theta_i\right)\right\}_{i=1}^m
\right).
 \end{align*}
 \begin{align*}
\centr&\left(U,V\right)\!=\! \\
&\sum_{i=1}^m {\left(\left(s x_i \right)^2+\left(s y_i\right)^2+\left(\left(1-s\right)\theta_i\right)^2\right)} - \\
&\frac{\left( \sum_{i=1}^m s x_i\right)^2 + \left(\sum_{i=1}^m s y_i\right)^2 + \left(\sum_{i=1}^m \left(1-s\right) \theta_i\right)^2 }{m}.
\end{align*}
Note that for the case of $s=1$ (translation only) the metrics are identical to those discussed in \cref{sec:metrics}.

\subsection{Experimental results}

\begin{figure*}
    \centering
	
	\subfloat[\fcap Scenario]
	{
		\includegraphics[width=0.33\textwidth]{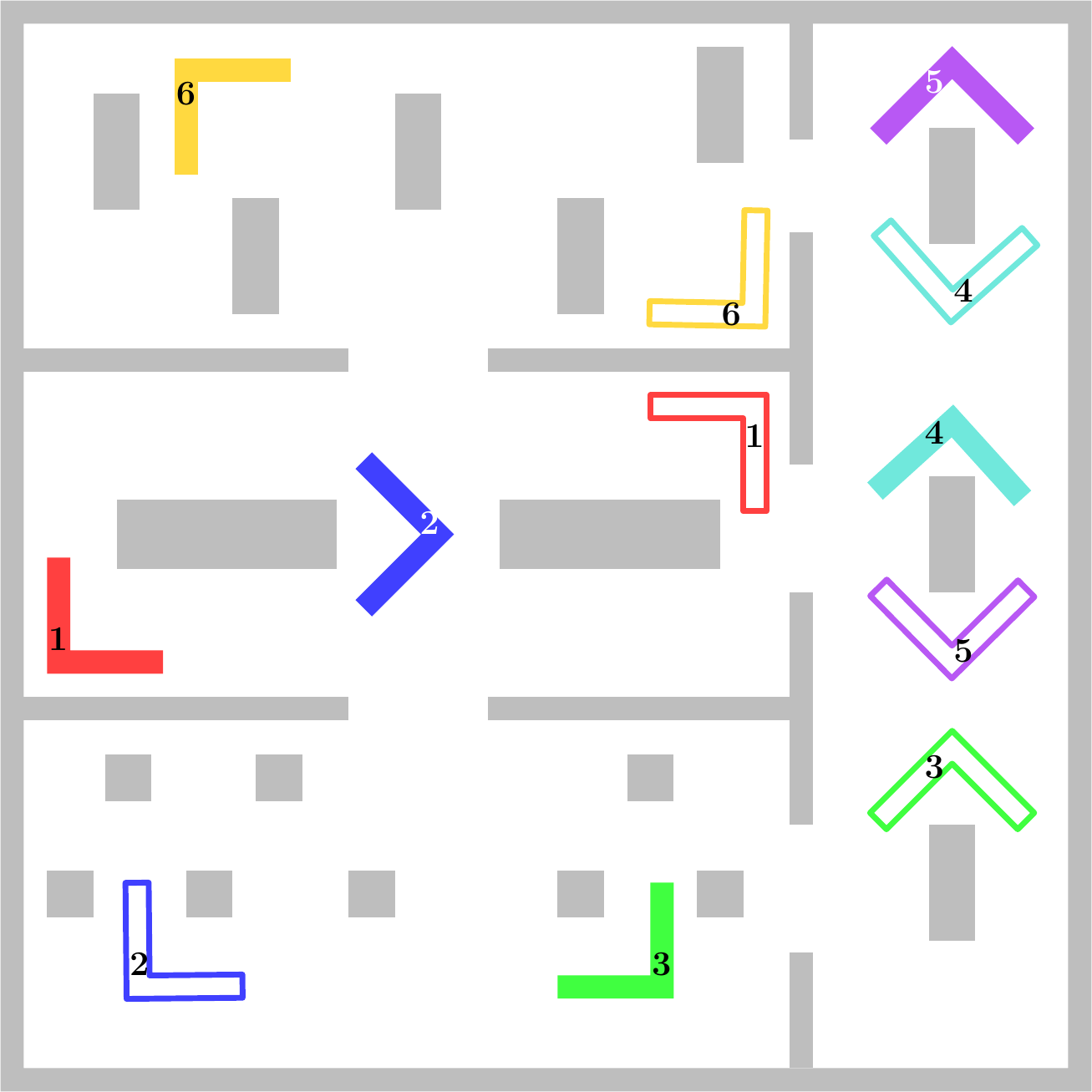}
		\label{subfig:lgrid_start}
	}
	\quad
	\subfloat[\fcap Effectiveness of metrics]
	{
		\includegraphics[width=0.6\textwidth, clip, trim=150pt 55pt 170pt 50pt]{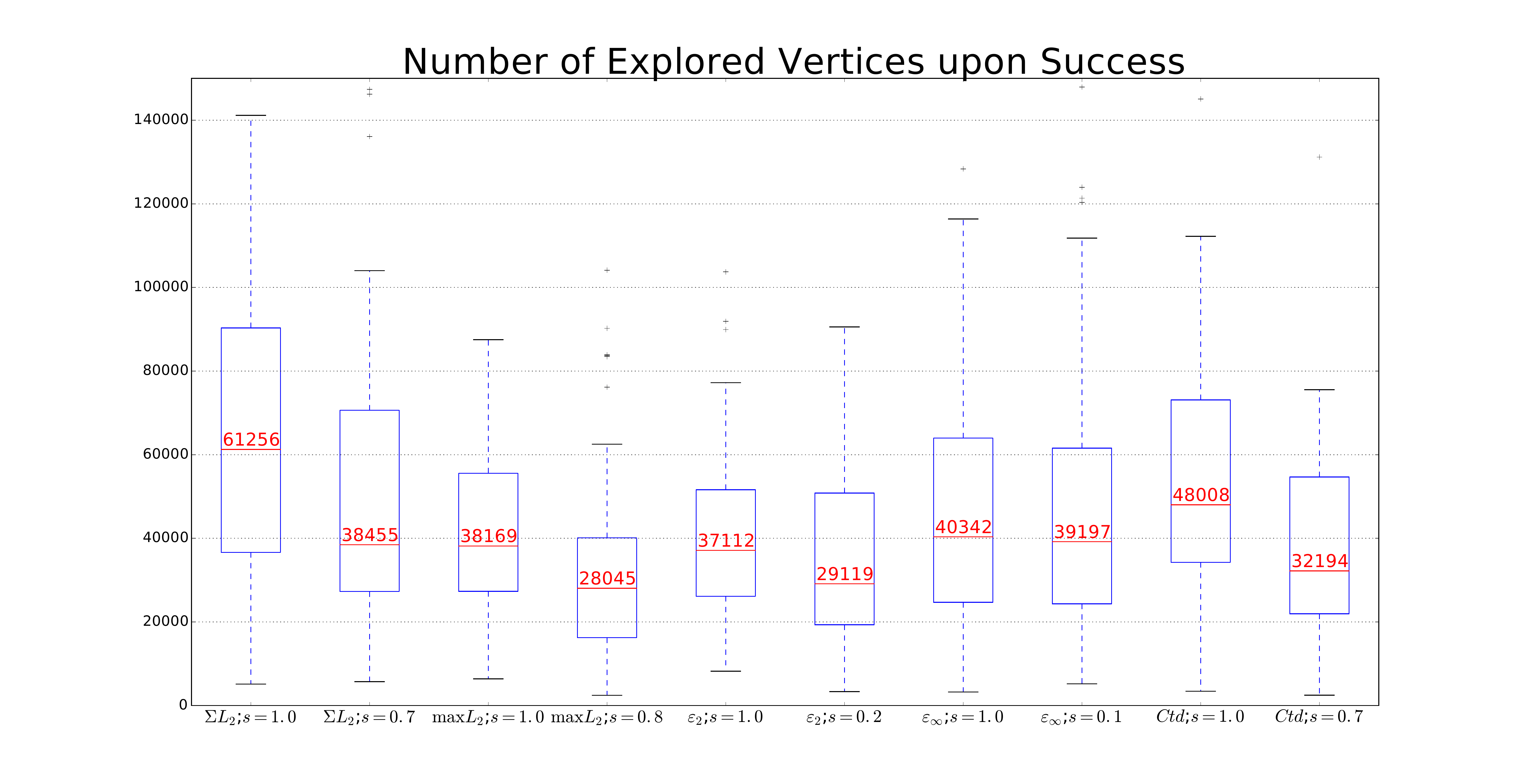}
		\label{subfig:lgrid_effectiveness}
	}
	
    \caption{\fcap A scenario with $m=6$ rotating robots in the plane. The robots are L-shaped and must rotate to arrive from the start configurations to the goal configurations.
      (a)~Start and goal configurations are drawn in solid and empty
      L-shaped, respectively.
      (b)~Number of expanded vertices when a solution is found. 
      The red labels are the median value.
      The experiment is repeated 50 times per metric and weight parameter~$s$.
      Effectiveness is expressed by low number of expanded vertices.}
    \label{fig:lgrid}
\end{figure*}

We use the scenario depicted in \cref{subfig:lgrid_start}. The robots in that scenario are L-shaped and allowed to rotate.

One difficulty that arises is deciding a proper value for the weight parameter $s$. Previous research has already addressed this problem~\citep{DBLP:journals/trob/AmatoBDJV00,DBLP:conf/icra/Kuffner04}. In our experiments we empirically chose the optimal value; we repeated the experiments 50 times for each value $s=0.1,0.2,\ldots,0.9$ and picked the one that leads to the most effective planning.
Note that the optimal value may be different for different metrics.

\cref{subfig:lgrid_effectiveness} shows the effectiveness of planning with each metric when the rotation component is taken into account. For each metric we show the effectiveness for~$s=1$ and for the optimal value of~$s$.
It can be seen from the figure that the adaptation of the metrics for rotating robots
takes rotation into account in a reasonable way;
%works as expected;
the effectiveness improves when we weigh in the rotation component of the robots. An exception is the case of~\epsinf, in which the improvement we get with the optimal value of~$s$ (when compared to~$s=1$) is not significant.
% the optimal value of~$s$ is~1.

%According to \cref{subfig:lgrid_effectiveness}, the most effective metrics are \Linf and \epst. Then, \epsinf and \centr, and finally, the least effective metric is \SLT.

According to \cref{subfig:lgrid_effectiveness}, the most effective metrics are \Linf and \epst. Then, \centr, and finally, the least effective metrics are \epsinf and \SLT.

\section{Conclusions and Future Work}
\label{sec:future}
\noindent Our work suggests that in order to effectively solve
MRMP using sampling-based planners one should employ tailored
multi-robot metrics, possibly side-by-side with more traditional
metrics. 

An immediate question is how to efficiently combine the benefits of different metrics. This resembles the idea of combining different heuristics in search-based algorithms~\citep{ASNHL16}. We propose to borrow ideas from this domain to address our problem. One approach may be to grow several trees, one for each metric. The trees can share states (such as SMHA*~\citep{ASNHL16}) and choosing which tree to grow at each point can be done in a dynamic fashion (see, e.g.,~\citep{PNAL15}).
%An alternative approach that we wish to consider is to grow a single tree.
%Here, we choose to use each metric according to some probability. The probability is initially uniform but is dynamically updated. Specifically, given a node~$q$, the algorithm will consider the set of nodes~$Q_{\text{near}}$ in the vicinity of $q$ and will use the prior information on the success or failure of each metric used to expand the nodes in~$Q_{\text{near}}$ to calculate the probability to use each metric.

This work has utilized three substructures and their
  combination in order to asses metrics. Of course there could be
  many more substructures, in particular larger, more elaborate
  ones, which would possibly improve our understanding of
  metrics. Thus, it would certainly be useful to automatically
  % construct these, by possibly
  % enumerating combinatorically all possible substructures, or 
  identifying these, possibly through a learning phase.

Metrics are relevant for other settings of MRMP, including those
involving moving
rigid bodies in 3D, and robots with differential constraints.
The proposed metrics and analysis tools can be extended to such
settings as well. 
% In fact, we already applied our ideas to simple settings with robots that can translate and rotate.

Another notable variant is the \emph{unlabeled} setting in which
all the robots are identical and interchangeable.  There are
similarity measures for unlabeled point sets that can be adapted
for MRMP
\citep{DBLP:journals/dcg/AltMWW88,DBLP:journals/pami/BelongieMP02,DBLP:conf/compgeom/EfratI96,hausdorff1927mengenlehre,DBLP:journals/dcg/HuttenlocherKS93}.
Unlabeled planning involves matching functions as well, which
have common properties with metrics but make the problem
considerably harder. We have began to explore the unlabeled case,
and have some promising initial results in this direction as
well.  A demonstration of our initial results for unlabeled disc
robots is provided in \ifx\arxiv\undefined the extended version
of this paper.  \else Appendix~\ref{supp:extensions}.  \fi

In this work we assessed metrics using RRT-style planners, and specifically dRRT (see \cref{sec:screening}). Although we do not believe
that our reported results are biased towards these specific types
of planners, it would be interesting to see whether the
conclusions can be reproduced for other planners, that operate
differently than RRT, e.g.,~PRM*,
RRT*~\citep{DBLP:journals/ijrr/KaramanF11} and
FMT*~\citep{DBLP:journals/ijrr/JansonSCP15}. This also leads to
the question of the effect metrics have on the quality of the
solution paths in MRMP.
%
%\aviel{Old:} Currently, we experimentally fine-tune the
%parameter $\tau$. It will be interesting to come up with a
%theoretical analysis of the choice of this parameter.  \aviel{I
%  suggest:} Currently, we have an intuitive guideline for the
%choice of $\tau$. It will be interesting to come up with a
%rigorous theoretical analysis of the choice of this
%parameter. \kiril{I feel that while this direction is important
%  there is no point at stating it unless we actually plan to
%  pursue it, or have some insight about it which we can share
%  with the readers.} \aviel{Ok. So we will remove this paragraph, and in case it bothers the reviewers we will put some effort in justifying the current choice.}

%%% Local Variables:
%%% mode: latex
%%% TeX-master: "../paper"
%%% End:

\section*{Funding}
This work has been supported in part by the Israel Science
Foundation (grant no.~825/15), by the Blavatnik Computer
Science Research Fund, and by the Blavatnik Interdisciplinary
Cyber Research Center at Tel Aviv University.
Kiril Solovey is supported by the Clore Israel Foundation.

\ifx\arxiv\undefined
\else
\begin{appendices}
 \section{Metrics Calculation}\noindent
\label{supp:calculation}
We consider the running time required for the calculation of each of
the five metrics described in the paper.  The calculation of \SLT,
\Linf and \centr is straightforward and requires $O\left(m\right)$ time, where $m$ is the number of robots.
However, the implementation of $\eps$-congruence-type metrics is a little
more intricate.

Given the joint configurations $U,V$, $\epsinf\left(U,V\right)$ can be
calculated by first finding a smallest enclosing square of the set of
$m$ points $\left\{ v_i - u_i \right\}_{i=1}^m$
(denoted by $\mathrm{MiniCube}\left( \left\{ v_i - u_i \right\}_{i=1}^m \right)$). Half of the square
edge length is the value of \epsinf. This yields a running time of
$O\left(m\right)$, using only subtractions and comparisons.

$\epst\left(U,V\right)$ can be calculated similarly, using the
smallest enclosing disc of the set
$\left\{ v_i - u_i \right\}_{i=1}^m$
(denoted by $\mathrm{MiniBall}\left( \left\{ v_i - u_i \right\}_{i=1}^m \right)$). The radius of the disc is the
value of \epst. The enclosing disc can be calculated in time
$O\left(m\right)$ \citep{DBLP:conf/esa/Gartner99,
  DBLP:journals/siamcomp/Megiddo83a}.

We now proceed to prove the correctness of the calculation for \epst.
The proof of correctness for \epsinf is analogous.  Recall that we are
given two sets of $m$ planar points
$U=\left(u_1,\ldots,u_m\right), V=\left(v_1,\ldots,v_m\right)$, and
our goal is to find the minimal value $R\in\R^+$ such that there
exists a transformation $T:\R^2\rightarrow\R^2$ that satisfies
$$
\max_{i=1,\ldots,m}
  L_2\left( T\left(u_i\right),v_i\right) \leq R.
$$
We denote the minimal $R$ by $R^*$. 
For each $i=1,\ldots,m$ we define $\delta_i=v_i-u_i$ and let $\Delta=\left\{\delta_1,\ldots,\delta_m\right\}$.

Let $R\in\R^+$.  We observe that $R \geq R^*$ if and only if there is
a point $p\in\R^2$ that lies in the intersection of the $m$ discs
with radius $R$ centered at $\delta_i$ for \mbox{$1\leq i \leq m$}.  The
point $p$ can be viewed as a translation $T$ for which
\mbox{$\max\limits_{i=1,\ldots,m} L_2\left(
    T\left(u_i\right),v_i\right) \leq R$}.
Hence, the problem reduces to finding the minimal value of $R$ such
that $m$ discs with radius $R$ centered at $\delta_i$ have a
nonempty intersection.  For a given $R$, the intersection is nonempty
if and only if there is a point $q\in \R^2$ that satisfies
\begin{equation}
\label{eq:smaller_R}
\max_{i=1,\ldots,m} L_2\left(q, \delta_i\right) \leq R.
\end{equation}
On the one hand, if there exists a point $q\in\R^2$ that satisfies \cref{eq:smaller_R}, then the disc with radius $R$ centered at $q$ is an enclosing disc for $\Delta$. On the other hand, the center of any enclosing disc with radius $R$ of $\Delta$ satisfies \cref{eq:smaller_R}.
In sum, for a given $R$, the intersection of the discs is nonempty if and only if there is an enclosing disc of $\Delta$ with radius $R$.

Hence, the radius of the minimal enclosing disc of $\Delta$ is the value of $\epst\left(U,V\right)$.

Note that it can be easily extended to higher dimensions. In the above proof, one should replace $\R^2$ with $\R^d$ and ``disc'' with ``ball''. It can also be extended to metrics other than $L_2$ (as is the case for \epsinf).

The running times of \SLT, \Linf, \epsinf and \centr are almost
identical.  However, in practice the constant in the running time of \epst is
non-negligible.  Although there is a fundamental similarity between
\epst and \epsinf, when taking the running time into account, \epsinf
has a slight advantage since it can be computed more
quickly.\vspace{5pt}

\section{Visualization Tool}\noindent
\label{supp:visualization}
In the initial phases of our study we used a visualization tool for illustrating the progress of the planner for different metrics, as more samples are added.
The tool is based on \pyplot~\citep{DBLP:journals/cse/Hunter07}. It is  used to generate videos that show the tree expansion process.
Recall that each iteration of an RRT-style algorithm proceeds in the following fashion:
\begin{enumerate*}[(a)]
\item \label{phase:rnd} sampling a random configuration $V$ from the joint \Cs
\item \label{phase:nn} finding its nearest neighbor $U$ in the tree
\item \label{phase:str} steering from $U$ in the direction of $V$, to obtain a configuration $W$
\item \label{phase:lp} calling the local planner to check the motion between $U$ and $W$ and adding $W$ to the tree in case the motion is free.
\end{enumerate*}
The tool is used to visualize all these steps.

We provide videos that demonstrate the simulation of the process. See
\cref{fig:visual_demo} for a screenshot and basic explanation of the
videos.

\begin{figure}[ht]
\centering
\includegraphics[width=0.45\textwidth]{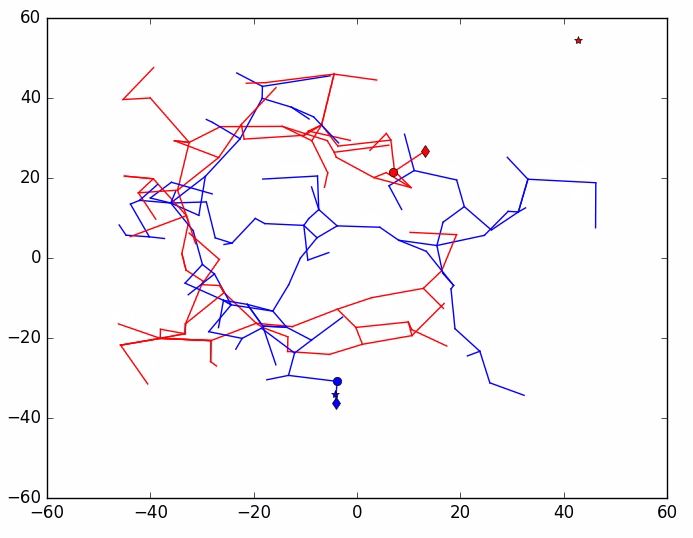}

\caption{A screenshot of the visualization-tool output. Full-size
  videos are available
  at~\url{https://www.youtube.com/channel/UCVBp--RJj7l4q-sDBA_rTbg}.
  Each robot is represented by a different color.
  In this scenario there are two robots ($m=2$), drawn in red and blue.
  The
  randomly-sampled configurations are drawn with stars.  The chosen
  configurations from the tree (nearest neighbor of the random
  configuration) are drawn with circles.  The configurations to steer to
  are drawn with diamonds.  The drawn edges are the tree edges
  projected onto each robot configuration space.  Sometimes it is more
  convenient to split the figure so that each robot has its own axis
  (refer to the youtube channel for such videos).  In order to reduce
  video time we omitted iterations in which the planner fails to
  expand the tree.}
\label{fig:visual_demo}
\end{figure}

As mentioned in the paper, one of the first type of metrics we have tested are
metrics that have high correlation with the failure rate of the local planner. 
We denote such a metric by \emph{CPM (Closest Point Metric)}.
The idea behind CPM is to calculate the closest point along the paths of any pair of robots, and accumulate all such closest points in order to predict how likely is the local planner to fail.
We compared CPM against the traditional \SLT in a scenario cluttered with random obstacles that involve two translating robots.
The visualization that illustrates the tree growth process for each metric is available at~\url{https://www.youtube.com/playlist?list=PLQFVBs-JqK1JbaJbliRtN6Y4qCZm6fOSs}.
It is noticeable from the video that CPM causes the planner to explore
configurations in which the robots are far away from each other, further causing them to be near the workspace boundary
The analogues for the single-robot setting are configurations in which the robot is located far from the obstacles.
Although it might be a desirable property for the single-robot setting, it raises difficulties for solving MRMP problems, since it is usually necessary to explore configurations in which the robots are not located near the workspace boundary.

%\pagebreak % to avoid splitting a link over a pagebreak
Videos used for visualizing the planner for the Tunnel scenario are available at~\url{https://www.youtube.com/playlist?list=PLQFVBs-JqK1Jyv-6Kc1ofDVmFuHU48jlQ}.
The videos illustrate the growth process of the tree until it contains 500 vertices.
One analysis tool that we describe in the paper is to count the number of distinct explored equivalence classes.
We show in the paper that the novel metrics that we propose cause the planner to explore more equivalence classes when compared to the standard metrics.
This phenomenon can be noticed in the videos.
For example, let us focus only on the order of the robots that lie in the upper ``arm'' of the workspace.
It can be observed that for \SLT (\url{https://youtu.be/8GBl6C9xxm8}), in most of the configurations, the topmost robot is the yellow robot, then the green robot and after them is the blue robot.
However, for \epst (\url{https://youtu.be/M_3b7J6aabA}), the order of the robots that lie in the upper arm is much more diverse. There are configurations in which the three topmost robots are the yellow, green and blue, while in other configurations the three topmost are the yellow, purple and cyan, and there are configurations in which the order begins with yellow, green and purple.
When the number of vertices goes up, the phenomenon becomes more extreme, as we show in the paper.

\section{Extensions} \noindent
\label{supp:extensions}
As noted in the paper, our methods can be extended to other settings
of MRMP. We already made some initial progress
% on settings where the robots are allowed to translate {\it and rotate} and
 on settings in which the robots are interchangeable with each other, i.e., the
so-called {\it unlabeled} case.

Animations of paths for the
% extended settings
unlabeled setting
are available
% at~\url{https://www.youtube.com/playlist?list=PLQFVBs-JqK1IhMLS9f9x4SIlrqPBfU4Kt}.
at~\url{https://www.youtube.com/playlist?list=PLNHY_VaTYKopM2JDpVMrWoyIowB4gXe__}.
We note that the work on the unlabeled setting is preliminary.

%%% Local Variables:
%%% mode: latex
%%% TeX-master: t
%%% End:

\end{appendices}
\fi

%% Use plainnat to work nicely with natbib. 

%\bibliographystyle{abbrvnat}
%\bibliographystyle{plainnat}
\bibliographystyle{SageH}
\bibliography{tex_references}

\end{document}